\documentclass[sigconf,nonacm]{acmart}

\usepackage{booktabs} %

\setcitestyle{nosort,square} %

\usepackage[ruled]{algorithm2e} %

\usepackage{color,colortbl}
\usepackage{tabularx}
\usepackage{makecell}
\usepackage{adjustbox}
\usepackage{multirow}
\usepackage{hyperref}       %
\usepackage{url}            %
\usepackage{amsfonts}       %
\usepackage{nicefrac}       %
\usepackage{microtype}      %
\usepackage{amsmath}        %
\usepackage{array}
\usepackage{xspace}
\definecolor{gray}{rgb}{0.85,.85,0.85}
\newcommand{\NXN}{$n$$\times$$n$\xspace}
\newcommand{\BSET}{\{0,1\}\xspace}
\newcommand{\DDE}{\emph{Down-Discretization Encoder}\xspace}
\begin{document}
\title{Texture for Colors: Natural Representations of Colors Using Variable Bit-Depth Textures}

\author{Shumeet Baluja}
\affiliation{
  \institution{Google, Inc.}
  \city{San Diego}
  \country{USA}
  }

\begin{teaserfigure}
  \centering
  \includegraphics[width=7in,height=2.7in]{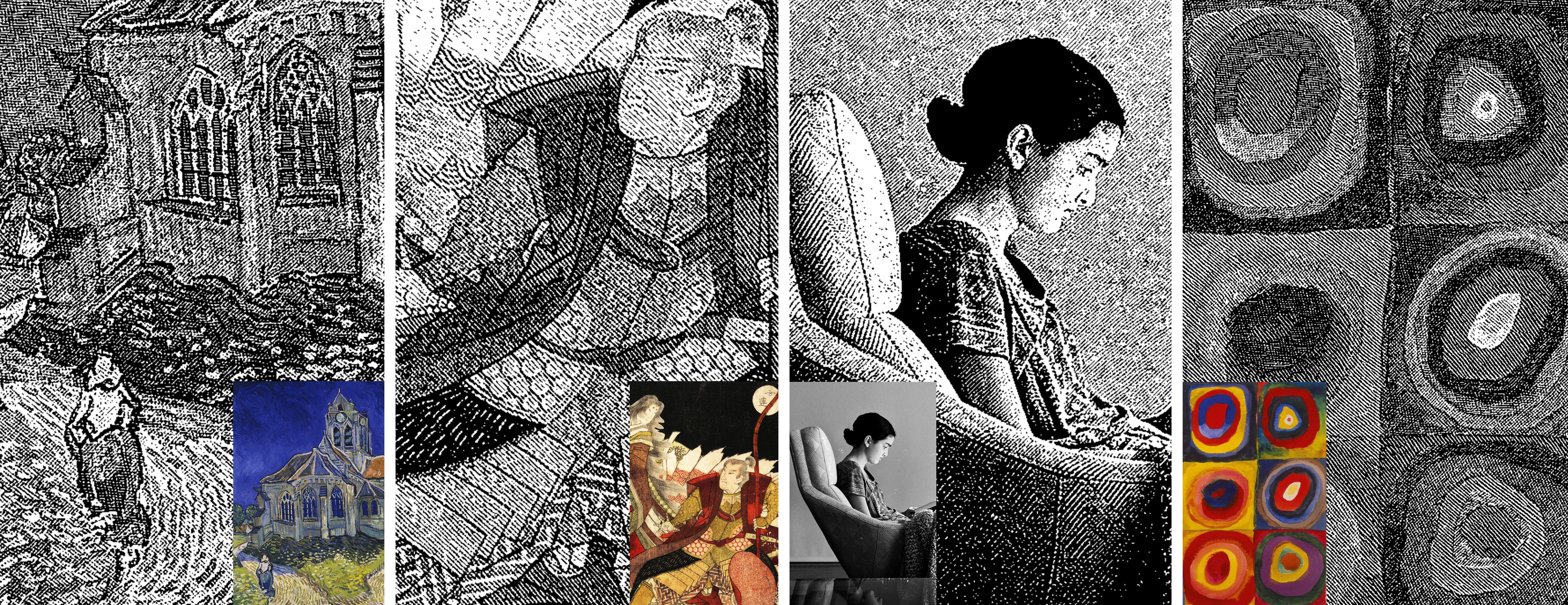}
  \caption{ Transformation of color images to textured, single-bit,
    images.  Similar colors are represented by similar textures,
    thereby allowing smooth gradients in the original to be visually smooth in the
    binary image as well.  The chosen textures accurately model the
    intensity variations and edges of the original image.
    Additionally, the textures are fully reversible; based solely on
    the binary texturized image, the full color original image can be
    recreated.  From left to right: \emph{The Church at Auvers}
    (1890) by Vincent van Gogh, \emph{The Warrior Miura-no-suke
    Confronting the Court Lady Tamamo-no-ma} (1820) by Yashima
    Gakutei, black and white photograph of a young girl reading
    (2020), and a portion of \emph{Color Study. Squares with Concentric Circles} (1913)
    Wassily Kandinksy.  }
  \vspace{0.2in}
\label{teaser}  
\end{teaserfigure}

\begin{abstract}

  Numerous methods have been proposed to transform color and grayscale
  images to their single bit-per-pixel binary counterparts.  Commonly,
  the goal is to enhance specific attributes of the original image to
  make it more amenable for analysis.  However, when the resulting
  binarized image is intended for human viewing, aesthetics must also
  be considered.  Binarization techniques, such as half-toning,
  stippling, and hatching, have been widely used for modeling the
  original image's intensity profile.  We present an automated method
  to transform an image to a set of binary textures that represent not
  only the intensities, but also the colors of the original.  The
  foundation of our method is information preservation: creating a set
  of textures that allows for the reconstruction of the original
  image's colors solely from the binarized representation.  We present
  techniques to ensure that the textures created are not visually
  distracting, preserve the intensity profile of the images, and are
  natural in that they map sets of colors that are perceptually
  similar to patterns that are similar.  The approach uses deep-neural
  networks and is entirely self-supervised; no examples of good vs. bad
  binarizations are required.  The system yields aesthetically
  pleasing binary images when tested on a variety of image sources.
  
\end{abstract}

\maketitle
\section{Introduction}

\begin{figure*}
  
  \centering
  \includegraphics[width=\textwidth]{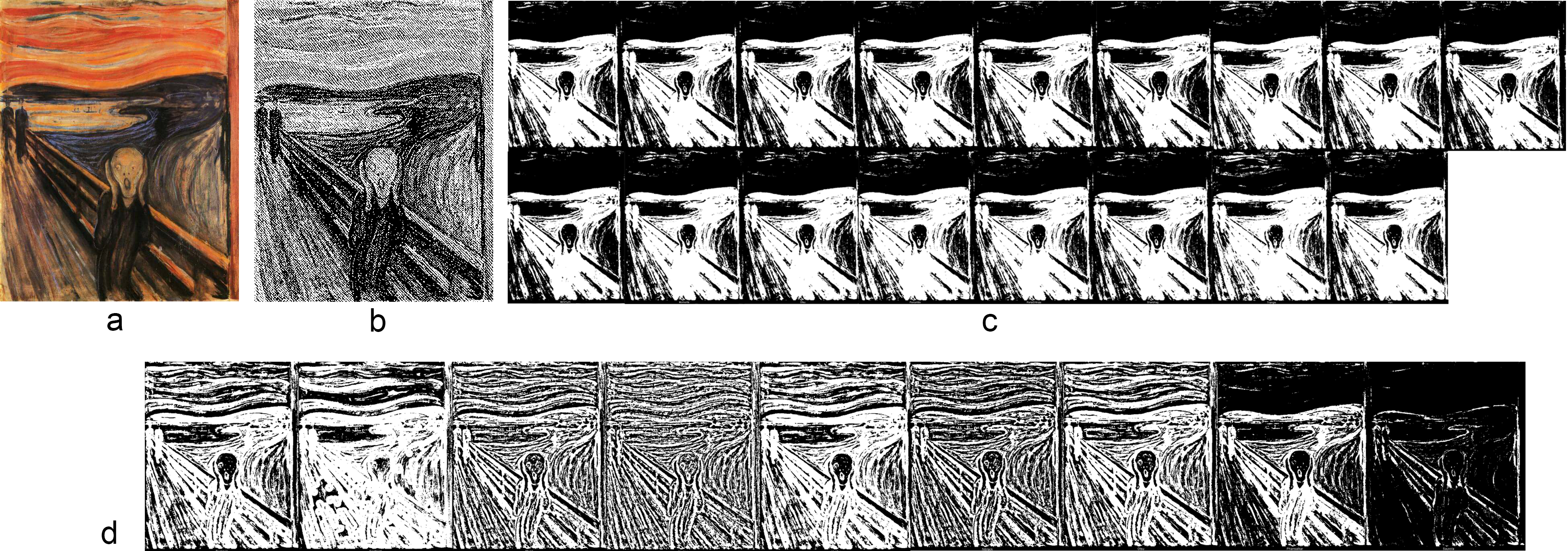}
    \caption{26 approaches for creating a binary image.
      (a) \emph{The Scream} by Edvard Munch (1893). 
      (b) Our approach.  (c) 17 global-thresholding methods (shown small, no detail is preserved).
      (d) 9 local-techniques.}
    \label{fig:imagej}

\end{figure*}

For more than half of a century, numerous methods have been proposed
to transform images to their binarized counterparts --- where each
pixel is represented by a single on/off bit~\cite{earliest}.  At the
highest level, these approaches can be categorized into two sets:
those intended for human viewing and those intended as a
pre-processing step for one specific task.  When created for
human-viewing, the images created are typically more aesthetically
pleasing and maintain a broader set of the original image's attributes
at the expense of losing specific details useful for specialized
analysis.  Common techniques include stippling, hatching and
edge-based approaches.  Early examples of cross hatching in Western
art, where hatches and cross hatches were used to represent a scene's
most salient image colors and features can be traced back to the
Middle Ages~\cite{hatchingWiki}.  In modern usage, for example in
newspapers, a variety of stippling approaches are employed.  These are described
in the next section.

Beyond artistic interest, binarization is commonly used in a variety of
scenarios.  Popular electronic-book readers~\cite{eReader-survey},
such as Kindle~\cite{eReader-kindle}, Boox~\cite{eReader-boox}, and
Nook~\cite{eReader-nook}, as well as public digital
signage~\cite{e-ink,e-ink-com} all use e-ink with varying levels of
bit-depth.  Binary texture images are the basis for a large set of
subtractive fabrication processes. These include water-jet,
laser-cutting with thin materials (paper, cloth,
\emph{etc.})~\cite{subtractive-glowforge}, or popular home cutting
devices~\cite{subtractive-cricut}.

We introduce a method to create selectable-bit-rate images from color
or grayscale originals by automatically synthesizing a texture
representation of the image.  We concentrate on the hardest case of a
binary image with a single bit per pixel.  Nonetheless, the techniques
are trivially modified to address representations with any number of
bits.  Even with just a single bit, the full color image can be
reconstructed from the result.  Additionally, images that contain
colors that are similar in intensity/luminance, or any dimension that
is commonly used for setting thresholding points or for converting color to
grayscale, will also be represented with distinguishable textures.  In
contrast to recent deep-steganographic
techniques~\cite{hu2018novel,baluja2019hiding}, there are no low-order
bits to effectively hide the original's color information. Instead,
all the information must be encoded into the \emph{visible} features
-- the textures that are created; see Figure~\ref{teaser}.

Two fundamental challenges must be addressed to create an effective
system.  In the standard 3-channel 8-bit RGB representation,
there are $2^{24}$ different colors possible.  Beyond the difficulties
associated with the number of unique textures required to represent
each possibility, we cannot utilize a unique texture for 
every pixel, as each is given only 1-bit in the binarized
image.  Instead, our method selects which colors to represent based on
the prevalence of colors in the images we wish to model (\emph{e.g.}
natural images).  It also learns to adaptively combine spatially close
pixels in the original image to create a texture representative of
localized regions.

Second, we must assign the textures to the colors judiciously -- such
that the resulting binary image does not introduce extraneous
edges.  For this, it is important that colors that are visually
similar have visually similar textures.  We refer to this as a \emph{natural}
ordering of the textures; the textures seamlessly blend into each
other when smooth transitions of colors are present in the original
image. %

This paper makes five contributions.  First, we have created a system
that transforms any full color image into a reduced bit-depth gray or
pure black and white representation.  The approach does not require
iterative search or relaxation techniques; a single forward pass
through a deep neural network is used.  Second, we demonstrate how our
method achieves natural textures such that visually similar colors are
represented with similar textures.  Third, within the learning
framework of deep neural networks, we present both novel architectures
and novel error functions to help capture aesthetic preferences.
Fourth, we demonstrate how we can recover the original colors from the
binary textures. Fifth, we briefly note how we alleviate some of the
biases in machine learning systems related to skin color by working
towards more accurate accounting for complexion differences and color
representation.

\section{Related Work}
\label{related}

\begin{figure*}
  \includegraphics[width=.99\textwidth]{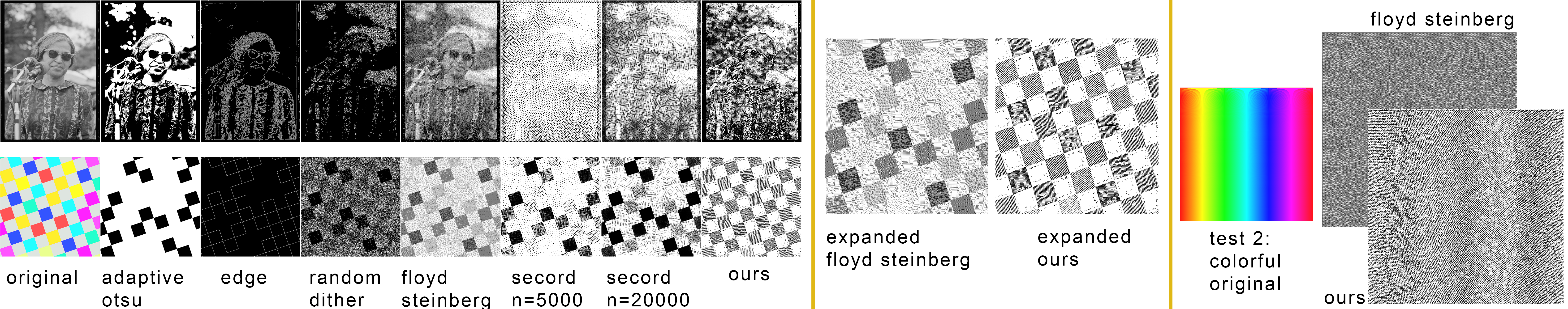}\\

    \caption{Image binarization intended for 
      human viewing.  (left) Original Image, Otsu Adaptive,
      edge detection~\cite{opencv}, random
      dithering, Floyd-Steinberg (FSD) Dithering~\cite{pillow}, Secord Voronoi with point sizes from 1 to
      20 (5000 points), Secord Voronoi with 20,000~\cite{rougier2017},
      and our method.  (middle)  Zoomed images of tilted square image
      binarized with FSD and our method.  FSD cannot
      distinguish similar intensity squares, despite different colors.
      Our method can capture them since we operate on full color images.
      (right) All colors are lost
      with FSD and methods that use only a single channel.  In
      contrast, our approach considers all channels.  }

    \label{fig:stippling}
\end{figure*}

There is a wealth of literature on approaches to
binarize/threshold images; many survey articles provide a
comprehensive history
\cite{sezgin2004survey,rasband1997imagej,surv_chaki2014comprehensive,surv_chang2006survey,surv_sahoo1988survey,surv_stathis2008evaluation}.
In Figure~\ref{fig:imagej}c, examples of 17 \emph{global}
thresholding methods are shown.  Global thresholding approaches
compute statistics from the entire image to decide where to place a
single threshold to create the binary cutoff.  %
Shown are 
two versions of
Huang fuzzy thresholding~\cite{huang1995image}, Prewitt's method
with bimodal histograms~\cite{prewitt1966analysis},  IsoData~\cite{ridler1978picture}, Minimum Cross
Entropy~\cite{li1998iterative}, MaxEntropy~\cite{glasbey1993analysis},
MinError~\cite{kittler1986minimum}, Prewitt's method with
histogram smoothing~\cite{prewitt1966analysis}, Moment
preservation~\cite{tsai1985moment}, Otsu~\cite{otsu1979},
foreground/background splitting assuming 50\% of
each~\cite{doyle1962operations},
Renyi Entropy~\cite{kapur1985new}, Shanbhag~\cite{shanbhag1994utilization},
Triangle Method~\cite{zack1977automatic}, and Yen's method for
multilevel thresholding~\cite{yen1995new}.  Methods were computed using 
the \emph{ImageJ}
toolkit~\cite{rasband1997imagej,schneider2012nih,abramoff2004image}.  As noted by Lee et al.~\cite{lee1990comparative}, 
global thresholding works well when the pixel
intensities in the image form a bimodal histogram and the object of interest is large in comparison to the image size.
However, issues
arise when the object of interest is small compared to the size of the
image, when there is low contrast between the foreground and
background pixels, when the image, or object of interest, has
non-uniform illumination or the image is degraded with significant
noise.  As can be seen in Figure~\ref{fig:imagej}c, the
majority of the details have been removed in all of the binarized
results.

In Figure~\ref{fig:imagej}d, results from nine \emph{local}
thresholding techniques are presented. These methods compute a cutoff
threshold based on the intensity statistics of surrounding pixels.
The results include the Bernsen method~\cite{bernsen1986dynamic},
local toggle contrast~\cite{soille2013morphological}, mean, median and
mid-gray thresholding based on local
statistics~\cite{adaptiveThreshold2003},
Niblack~\cite{niblack1986introduction}, localized
Otsu~\cite{otsu1979}, Phansalkar~\cite{phansalkar2011adaptive}, and
Sauvola~\cite{sauvola2000adaptive}.  %
The local approaches outperform the global by maintaining more
details and shading and intensity variations.

In contrast to the task-specific approaches above, the next
set of techniques are more often used for creating binarizations
suitable for human viewing, for example in newspapers.  Common techniques in
this class of approaches include stippling
methods~\cite{funkhouser2000} such as dithering, half-toning and
stroke-based approaches that include hatching and edge detection,
see Figure~\ref{fig:stippling}.  Here, the distribution of dots is
vital to the perception of the underlying image.  The canonical method
is based on error diffusion~\cite{FloydSteinberg1976}.  Intensity can
also be modeled with variable size dots.  Secord's approach based on
Voronoi regions has yielded compelling
results~\cite{secord2002weighted,rougier2017}; samples are shown in
Figure~\ref{fig:stippling}.  Hatching and cross-hatching use variable
length and thickness lines oriented in parallel, crossing, or
contoured orientations.  Prominent automated approaches have tackled
shading and contours for complex
objects~\cite{zander2004high,praun2001real,kalogerakis2012learning}.
Edge-detection also falls into this category of line-based
representations of images.

We have observed Floyd-Steinberg dithering and methods such as the
Linde-Buzo-Gray preserve details from the
original~\cite{deussen2017weighted}.  However, there is an underlying
limitation to these approaches: they must operate on grayscale images
when the desired goal is a single bit-per-pixel (1-bpp) image.  If the
original has colors, the first step is to convert it into a single
channel~\cite{hellandDithering}.  All standard conversion operations
lose information that will yield disparate colors indistinguishable.
In contrast, with our system, all three channels are considered; the image is \emph{not} first
converted to grayscale, allowing us to encode and recover the colors
solely from the binary image (Figure~\ref{fig:stippling}-right).

The closest work to ours in terms of motivation is
\emph{Color2Gray}~\cite{gooch2005color2gray}. Color2Gray attempted to
handle the isoluminant variations that were not preserved with
standard color to grayscale conversions through explicitly analyzing
chrominance and luminance differences.  Like
~\cite{gooch2005color2gray}, we handle isoluminance; however, we
employ an information-preservation perspective and dramatically reduce
bit rates. Similar problems have arisen in numerous specialized
applications~\cite{qu2008richness} that require converting photographs
to stylized bi-tonal renderings.  In terms of the approach taken, the
closest work is~\cite{xia2018invertible}.  They use a CNN to convert
color to grayscale and back; the work parallels the deep-steganography
mentioned earlier.  Their architecture is a subcomponent of the one
presented here.  For example, to achieve 1bpp while maintaining smooth
transitions in colors, their method requires augmentation with both
the novel loss functions and the architectural mechanisms that will be
presented.

A note about the relationship of this work to style
transfer~\cite{gatys2016image,gatys2015neural},
GANs~\cite{makhzani2015adversarial,goodfellow2014generative}, and
Image-to-Image
Translation~\cite{isola2017image,liu2017unsupervised,zhu2017unpaired}.
Our initial approaches to synthesizing a binary image from a full
color image was to use GANs within the unsupervised image-to-image
translation process and numerous style-transfer approaches.  Though
good results were sometimes obtained, there were severe drawbacks.
With style-transfer, the same color on different original images were
not consistently mapped to the same textures -- an intuitive
expectation.  Second, in order to ensure that the result was 1-bpp,
the majority of the contributions presented here, the architectures and
loss functions, were needed to supplement all the approaches.  Finally, the GAN training process was
significantly less stable than using explicit information-preservation
as the basis of our objective function.  The final system described
here did not require an adversarial teacher.  This yielded a simpler and 
more stable to train system.  It also provided consistency in color
mappings across images, and generated at least equal, and most often superior,
results.

\begin{figure*}
  \includegraphics[height=1.60in]{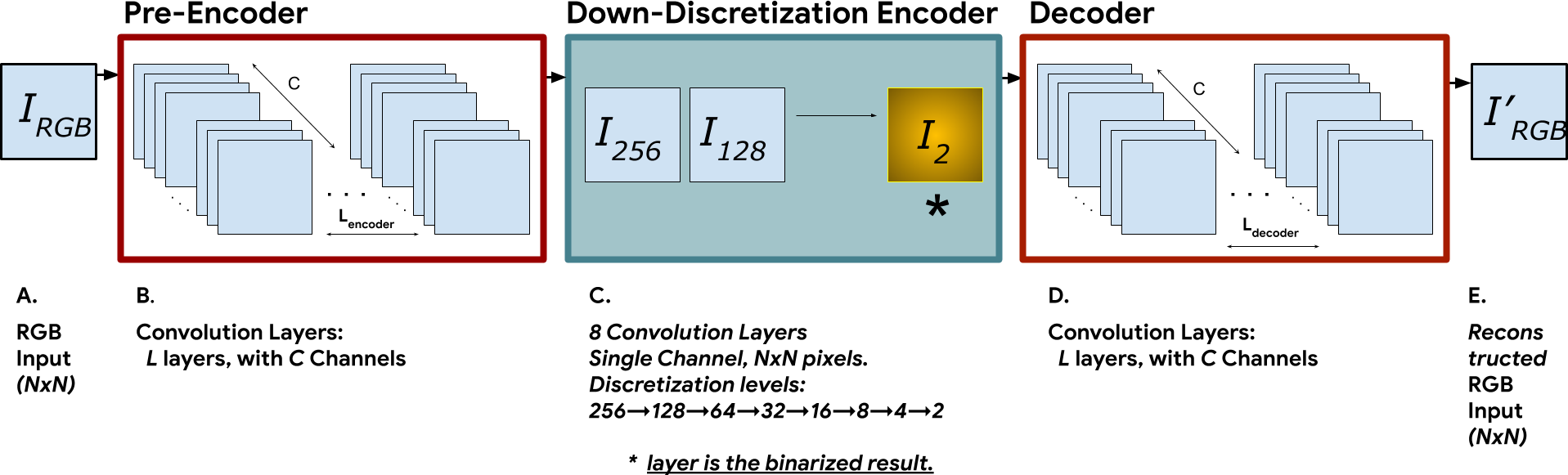}~~~~~
  \:\:\:\:\:  \:\:\:\:\:  \:\:\:\:\:  \:\:\:\:\:
  \includegraphics[height=1.60in]{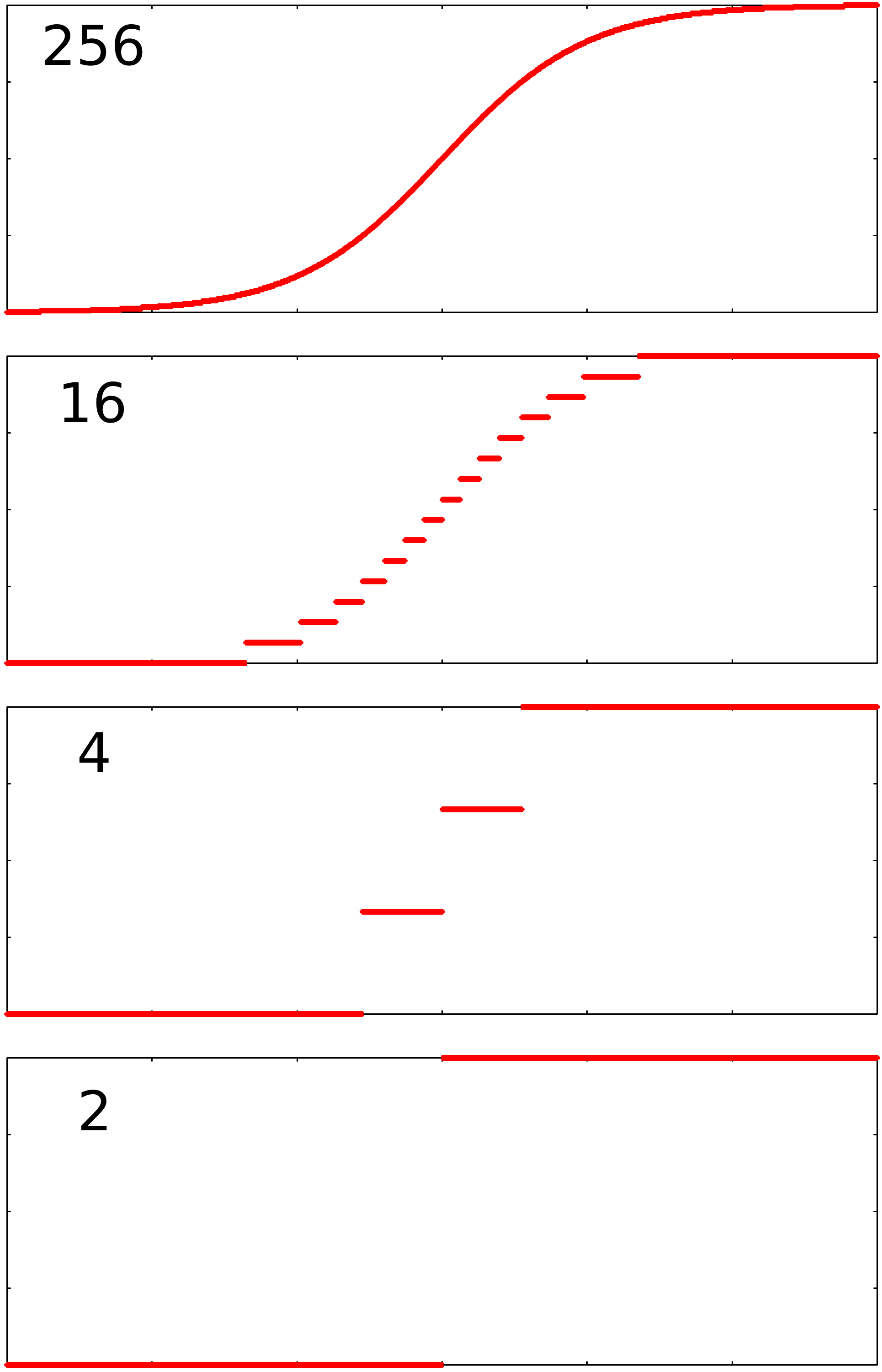}

  \caption{(left) A fully convolutional network architecture to transform an \NXN~24-bpp RGB
      image into a \NXN~1-bpp binary image.  We cast the problem
      as a reformulation of an auto-encoder network with explicit
      constraints on the bottleneck's representation.  Note
      that the entire decoder module is used exclusively in training;
      we need it to compute the error that is used to train the
      system: $L_2 (I_{RGB}-I'_{RGB})$.  The actual image that will be
      shown to the user, the binarized image, is marked with an
      asterisk (*).  (right) The discretized $tanh$ activation functions for 4 of the 8
      layers in the \DDE (256,16,4,2)%
      }
    \label{fig:netArch}
\end{figure*}

\section{Learning Textures}
\label{learning}

In our work, we attempt to represent unconstrained complex scenes
(full photographs, paintings, artwork) as binary images.  In
designing our approach, the seven most salient desiderata are:
\begin{enumerate}
    \item \emph{Different perceived colors should have different texture
      representations.}  The representation chosen for a color should not
      solely depend on any single dimension (\emph{e.g.} intensity).
    \item \emph{Colors that are visually similar are represented by similar
    textures.}
    \item \emph{The same color in different images should be
    represented similarly.}  This allows the reconstruction of
      the original's colors and matches intuitive expectations.
    \item \emph{The textures introduced to represent colors should not be
      visually distracting.}  The prominent edges in the original should
      be the only ones visible under normal viewing magnification.
    \item \emph{The approach should be extendable to variable bit-rates.}
       Some e-ink displays have a 2-level
      display, while other widely deployed e-readers, such as 
      Kindle, have 16~\cite{eReader-kindle}.
    \item \emph{In addition to representing the colors of the original image,
    images without color (\emph{e.g.} grayscale originals) should also work.}
    \item \emph{We should not have to manually create textured images 
      to train the system.}
\end{enumerate}

Given the desiderata described above, let us first consider the
simplest solution: mapping every color to a unique texture and simply
using a lookup table. To implement this approach, first, we would need
to decide which discretization of colors to consider, since, as
mentioned earlier, $2^{24}$ textures is unwieldy.
 Second, after a
discretization scheme is chosen, how should we measure similarity for
this task?  Should hue, intensity and saturation be weighted equally,
or should alternate perceptual metrics be used?  Third, we cannot
replace a single pixel with a full texture, therefore we must decide
on how to create regions to model together.  Will this involve static
blocks or adaptive region growing?  Fourth, how are transitions from
one color to the next handled in the binarized image?  If this is
handled poorly, edges will appear in the image. Colors that often
occur together must have compatible textures to avoid creating visual
discontinuities in the binarized image as a result of minor color
transitions in the original image.

Instead of attempting to manually find a mapping from colors to
textures, we use a learning approach.  It is grounded in the following
observation: if the colors are represented uniquely in the transformed
image, then we should be able to invert this process, \emph{e.g.}
reconstruct the full color image from just the binary image. The
remaining desiderata are achieved through the specifics of the
training procedure as well as the novel loss functions that are
employed.

To begin, let us examine the well established principles of
auto-encoder neural
networks~\cite{kramer1991nonlinear,jiang1999image,larsen2015autoencoding,theis2017}.
An auto-encoder network is a self-supervised architecture to learn a
compressed representation of the input data.  The \emph{encoder}
portion of the auto-encoder learns to transform an input image to a
substantially reduced dimensionality (similarities to principal
components analysis are explored in~\cite{kramer1991nonlinear}). This
is accomplished by passing the input (an image) through a series of
non-linear transformations, including a low-dimensional ``bottleneck''
layer.  The \emph{decoder} portion of the auto-encoder manipulates the
internal compressed representation emitted by the bottleneck to
recreate the original image.

Through the backpropagation of errors between the original input and
its reconstruction after the decoding, the encoder learns to compress
information about the original image through the bottleneck. For images of resolution \NXN, a bottleneck
of size $m$ is used where $m \ll n\times n$.  Simultaneously, the decoder learns to use that representation to reconstruct the image.
Recently, reductions in transmission size competitive with
state-of-the-art image and video compression have been obtained
through auto-encoder based
approaches~\cite{toderici2016variable,minnen2018imagedependent}.

\begin{figure*}
  \centering
  \includegraphics[width=0.6\linewidth]{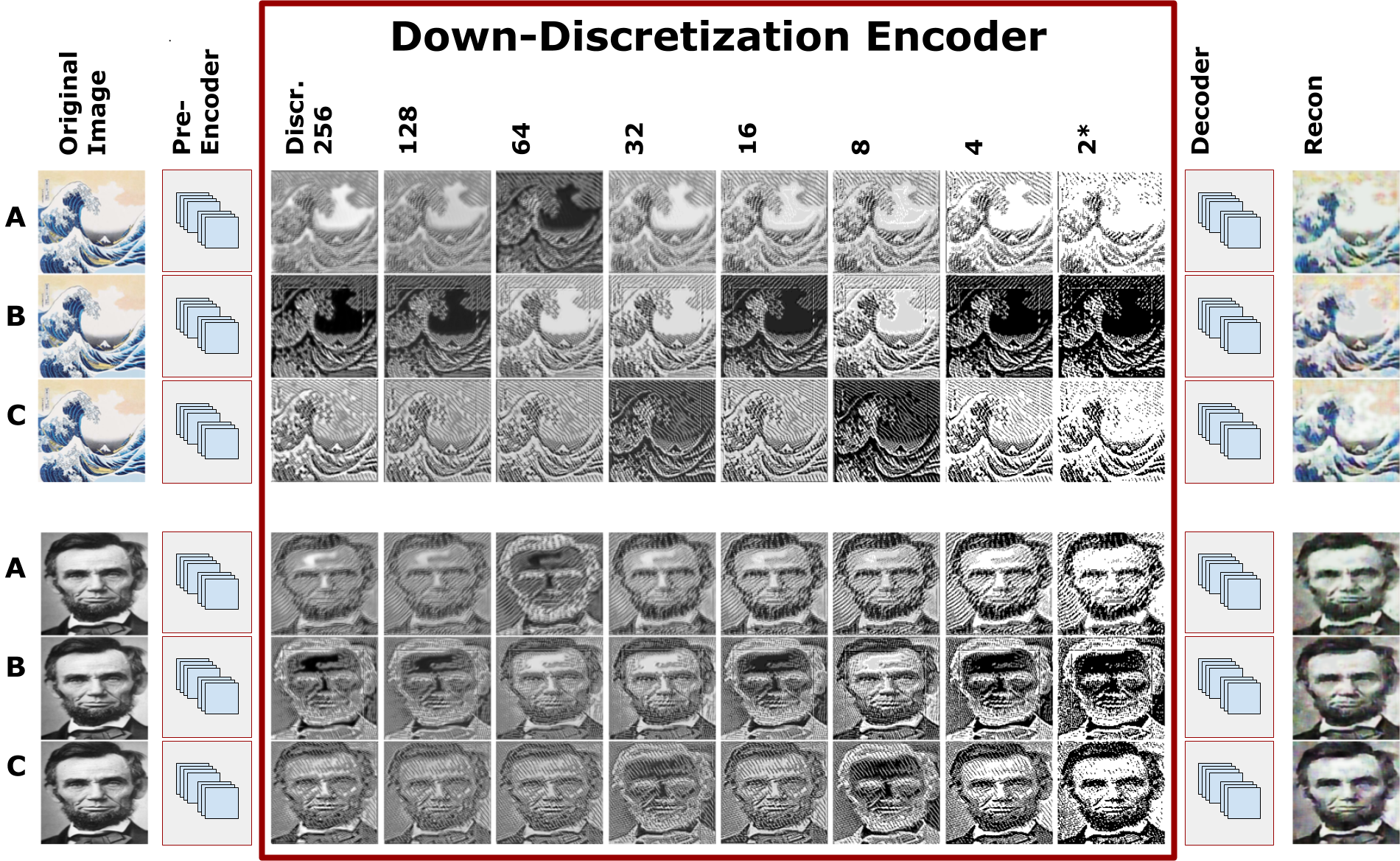}\\
  \: \: \:\\
  \begin{tabular}{ll}
    \shortstack{Enlarged \\Output\\ Layer (2*) \\ ~~ \\ ~~ \\ ~~ \\ ~~ \\ ~~ } &
       \includegraphics[height=0.9in]{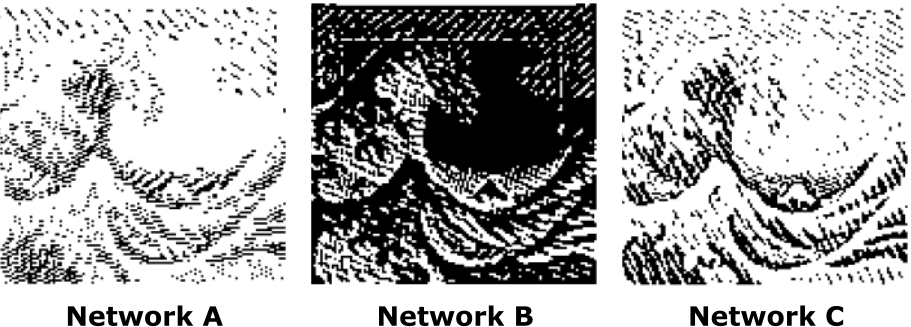}
  \end{tabular}
  
    \caption{(Top) The eight down-discretization steps shown for two
      images, \emph{The Great Wave off Kanagawa} (1829-1833) by
      Hokusai and a portrait of Abraham Lincoln.  The original image
      is leftmost (24 bit, RGB).  This is passed through the
      pre-encoder.  The next column, ``Discretized-256'', shows the
      output of the first Down-Discretization layer, in which each
      pixel can be 1 of 256 values (8bpp).  The next 7 layers reduce
      the image to binary by each removing 1 bit.  The actual output
      that the user is given is the last one in the \DDE (marked with
      (*)). All layers after the \DDE are only used for training.  The
      operations of three independently trained networks (A-C) are
      shown.  (Bottom) Enlarged versions of the binary images
      generated.  Note the variations resulting solely from the
      stochasticity in training between all three networks, including
      the inversion of intensities in Network B.}
    \label{fig:firstResults}

\end{figure*}

We use the same principle, but instead of reducing the
number of dimensions, we reduce the representational capacity of each
dimension.  For generating binary images, our equivalent to the
bottleneck layer keeps the same image dimensions as the original, \NXN; however, each
dimension can only represent two values: \BSET.  A network
architecture to implement this is shown in Figure~\ref{fig:netArch}.

Note that the reduction of colors to binary is accomplished within the
\DDE module (see Figure~\ref{fig:netArch}, Box C).  Here, the bits per
pixel in the image are successively lowered to 1-bit per pixel (bpp).  The input that
the \DDE module receives is an internal representation based on the
original 24bpp image.  The output of the module is the desired 1-bpp
image.

Figure~\ref{fig:firstResults} provides a visual explanation of the
transformations in the \DDE. The original image, in the leftmost
column, is a full 24-bpp image.  It is transformed by the Pre-Encoder
to an internal representation of $128\times$\NXN floating point
numbers.  This is used as the input to the \DDE.  The first
discretization layer receives this input and produces a single channel
\NXN ~8-bpp image.  This is fed into the next discretization layer
through a single (trained) convolution kernel and a single channel
7-bpp image is produced.  This process is repeated until there a 1-bpp
image.  Note that each layer's discretized-tanh activation introduces
a non-linear transformation, as shown in Figure~\ref{fig:netArch}.
The final layer's output from the \DDE is the binary image that is
shown to the end-user (marked with '*'). %

When training the system, the binary image emitted by the \DDE is
propagated through the decoder module.  The decoder reconstructs
the colors of the original image from the binarized representation.
The networks is trained to minimize the error between the
reconstruction and the original.  When successfully trained, the
textures in the binarized image uniquely identify the underlying
colors that generated the pattern.

Why should textures arise?  The key insight is that textures \emph{must}
arise in order to successfully reduce the reconstruction error.
Consider the alternative: an architecture in which the reconstruction
of each pixel was based on a single pixel of \BSET values.  The color
information required for reconstruction could not be encoded.  By
using convolution layers that allow nearby pixels to be considered,
groups of nearby pixels convey a region's color information.  However,
since we have constrained the system so that \emph{the only
information transmitted is through the visible binary image}, the
information encoding of nearby pixels takes the form of \emph{visible}
patterns; these patterns are precisely the textures desired.

For our first trial, we used the network shown in
Table~\ref{architecture}. We begin our experimentation by simply
minimizing the reconstruction error: $L_2(I_{RGB}-I'_{RGB})$.
ImageNet photographs (without labels) are used for
training~\cite{deng2009imagenet}; the inputs and outputs were
$128\times128$ pixels.  The networks were trained with Tensorflow,
using Adam optimization~\cite{abadi2016tensorflow,kingma2015adam} with
$LR=0.0001$, batch size=16 and uniform noise added to the inputs of
$\pm 5\%$.  To handle training in the presence of the discretizations
employed in the \DDE, we discretized the values from the units in all
of the forward passes (training and inference). However, in the backward
error propagation step of training, the discretizations were ignored
and the underlying function (tanh) was used instead; this is a
straight-through-estimator~\cite{hinton2012neural}.

\begin{table}
 \renewcommand{\arraystretch}{1.2}  
 \caption{Network architecture. %
   Each layer uses ($3\times3$) convolutions.\\(The final binarized output is marked with *)
 }
  \label{architecture} 
  \footnotesize
  \begin{tabular}{c|ccccc}
    & \rotatebox{0.00001}{input} 
    & \rotatebox{0.00001}{\makecell[l]{layers, \\channels\\per layer}}
    & \rotatebox{0.00001}{\makecell[l]{activation\\function}}
    & \rotatebox{0.00001}{\makecell[l]{activation\\ function\\ in last layer}}
    & \rotatebox{0.00001}{output}\\
    \hline
    \hline    
    \makecell{Pre\\Encoder}         & \makecell[c]{24 bit\\\NXN}  & 10, 128 & relu & relu &\makecell[c]{128 float\\\NXN} \\
    \hline    
    \makecell[c]{Down\\Discretization\\Encoder}  & \makecell[c]{128\\float\\\NXN}    & 8, 1 & \makecell[c]{tanh\\discrete$_{256}$-\\discrete$_{4}$} & \makecell[c]{tanh\\discrete$_{2}$} &
                         \textbf{\makecell[c]{1 bit\\\NXN(*)}}\\
    \hline    
    Decoder             & \makecell[c]{1 bit\\\NXN}     & 10, 128 & relu & tanh &  \makecell[c]{24 bit\\\NXN}\\
  \end{tabular}

\end{table}

\begin{figure}
  \footnotesize
  \begin{tabular}{ll}
      a. &
      \includegraphics[width=0.9\linewidth]{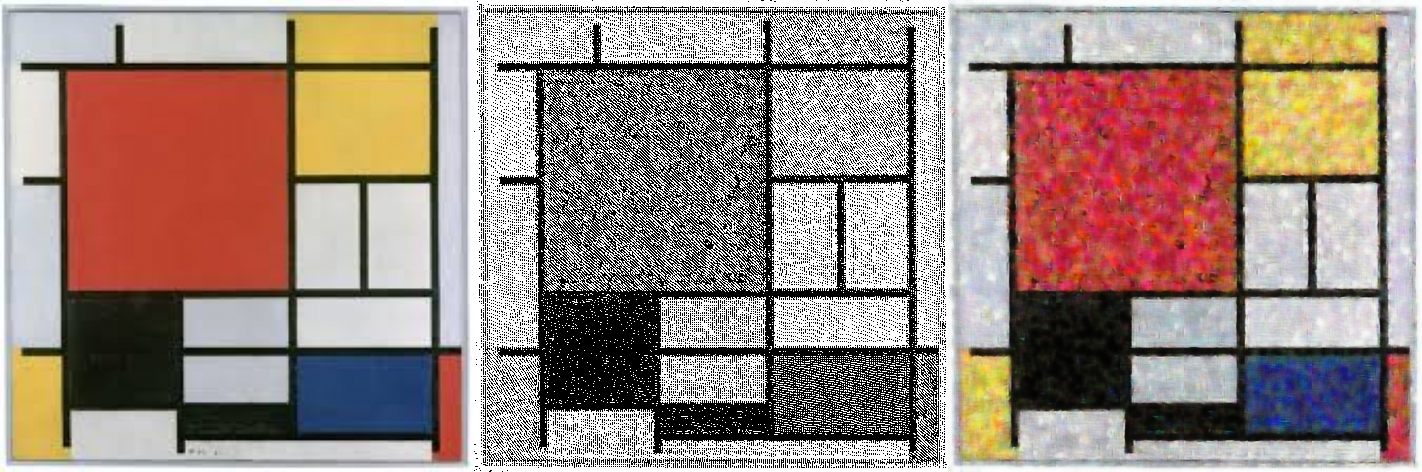}
      \\ b. &
      \includegraphics[width=0.9\linewidth]{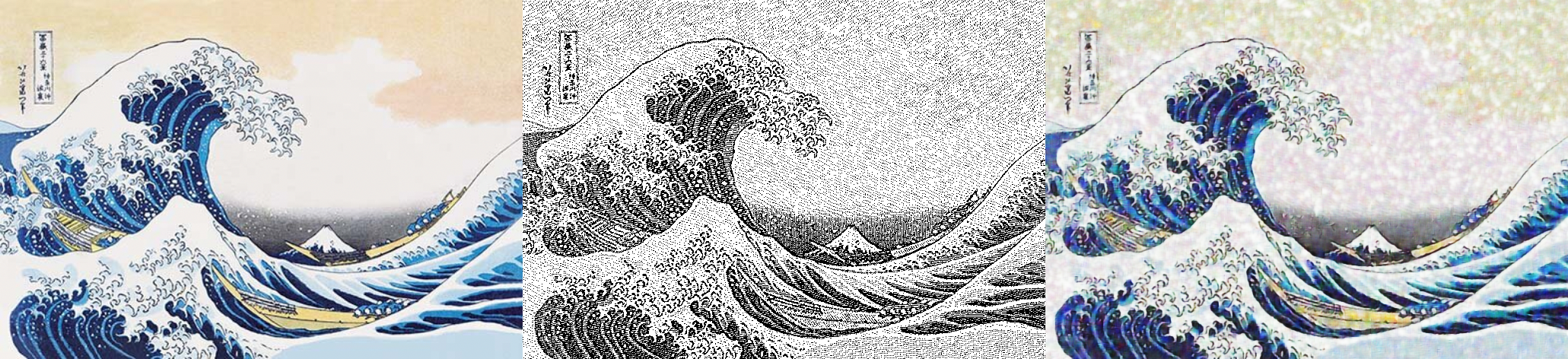}
      \\ c. &
      \includegraphics[width=0.9\linewidth]{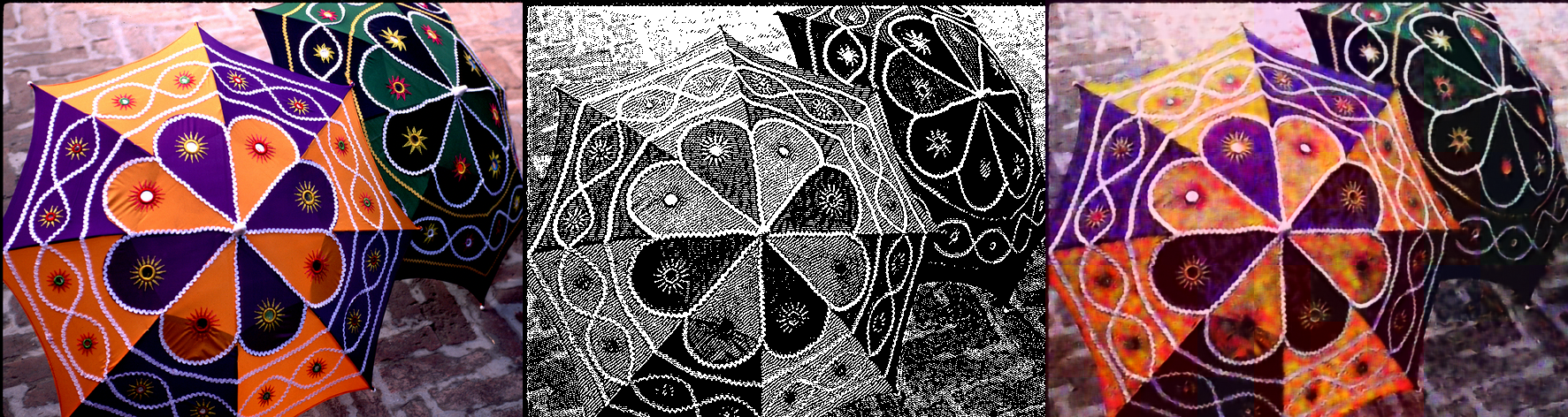}
      \\&
      \begin{tabular*}{\linewidth}{c @{\extracolsep{\fill}} ccc}
        original& \: binary& \:reconstructed
      \end{tabular*}
      ~\\
      ~\\
      ~\\      
      \\ d. &
      \includegraphics[width=0.9\linewidth]{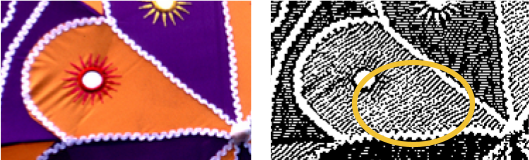}      
  \end{tabular}
    \caption{Samples from initial trial. In each triplet (a,b,c), the original (left),
      binary(middle), and reconstruction solely from the binary (right) 
      are shown. (a) Mondrian (1921): different colors
      are represented with different textures.  
       (b) Colorful umbrella photograph. (c) The wave's shape
      is clearly visible, but the white appears gray. (d) 
      A mistake from the umbrella photograph is shown enlarged: a drastic shift in
      textures does not correspond to a 
      large change in the original's colors. }
    \label{fig:firstResultsMoreSamples}
\end{figure}

Results of the initial tests were shown in
Figure~\ref{fig:firstResults} and are expanded upon in
Figure~\ref{fig:firstResultsMoreSamples}.  The binarized image produced by
network A and C (Figure~\ref{fig:firstResults}) appear reasonable, but
network B produces a color-inverted image.  This inversion occurs
because there is no information loss by inversion and therefore
reconstruction error (our sole measurement so far) is not affected.
Note that both networks A and C suffer from this in earlier stages of the
down-discretization procedure.  This is not just a problem of
inversions; rather, it occurs because there are no constraints for
matching the original's intensities in the \DDE.  We will address this
next, in Section~\ref{intensity}.

In the expanded results in Figure~\ref{fig:firstResultsMoreSamples},
note that the images' colors are successfully represented by
different textures.  The reconstructions, which are solely based on
the patterns in the binary image, recreate the colors well, though are
noisy -- indicating that some color information is not recovered.
Recall, however, that the reconstruction is \emph{not} the end product
in this task; it is never shown to users.  Its only purpose is to
compute the training loss.  An unwanted large discontinuity in the
binary textures is shown in Figure~\ref{fig:firstResultsMoreSamples}d.
This flaw occurs because visually similar colors in the original are
\emph{not} constrained to similar texture representations.  This can
cause the perception of edges where they do not exist in the original.
In Section~\ref{continuity}, we will describe how to improve this.

Though the intent of our procedure is to create textures for binarized
images, we note that the textures created are reversible.  They can be
converted back into colors (as shown in the triplets in
Figure~\ref{fig:firstResultsMoreSamples}a-c).  This occurs because the
reconstruction error is minimized in the training process.  We will
return to this in Section~\ref{conclusions}.

\subsection{Algorithmic Improvements}
\label{improvements}

Numerous extensions and improvements were attempted.  Those that had
the largest impact are described here.

\subsubsection{Relative Intensity Constraints}
\label{intensity}

To avoid the mistake shown in Figure~\ref{fig:firstResults}, in which dark regions in the original may be
represented with bright textures, intensity constraints are introduced. 

We need to ensure that the relative brightness of region$_j$
vs. region$_k$ in the original image, $I$, matches the relative
brightness of region$_j$ vs. region$_k$ in the binary image $I_2$.  We
divide $I$ into $r$ non-overlapping equally sized regions, and compute
each's average intensity.  The result is reshaped into a column
vector, $A$, of length $r^2$.  Next, we compute the difference in
intensities for each pair of regions, $A' = A \cdot{\bf 1}^T - {\bf 1} \cdot A^T$  (the {\bf 1} column vector is of length $r^2$).  The
differences are scaled to [-1.0,1.0].%

The corresponding matrix, $B''$, is computed for $I_2$.  With this, we
compute the magnitude of the relative differences, $|A''-B''|$.  The
lower this value, the more the differences in region intensities
\emph{within} the original images match the region differences
\emph{within} the binarized image.  This measurement
(Equation~\ref{intensityloss}) is used to supplemental 
the $L_2$-reconstruction loss described earlier.  In our tests, we set
$r=8$.

\begin{equation}
  \label{intensityloss}  
  \begin{split}  
  L_{relative\_intensity}=\\&\hspace{-0.5in}|tanh (A \cdot{\bf 1}^T - {\bf 1} \cdot A^T) - tanh (B \cdot{\bf 1}^T - {\bf 1} \cdot B^T)|
  \end{split}
 \end{equation}

We note that this step is particularly important to address the
sensitive topic of hidden biases in the training and evaluation of
machine learning methods.  Without this step, varying skin-tones were
mapped to textures that appeared similar.  There was little
distinction between paler and darker complexions.  Adding this step
significantly improved the distinctions in the resulting binary images
between light and dark skin tones.

\subsubsection{Color Continuity Constraints}
\label{continuity}

To avoid the mistake
shown in Figure~\ref{fig:firstResultsMoreSamples}, we need to ensure
that similar colors are represented by compatible textures.  Let us
consider a family of modification operators, $M(I)$, that perturb the
input $I$, such that the colors in $I$ are shifted by a small amount.
If the modification operators we consider yield images that appears
similar in terms of color, then by ensuring that the textures used for
the colors in $I$ match the textures for $M(I)$, we ensure that
similar colors are represented similarly, see
Equation~\ref{continuityEqn}.  The network
transforms the input $I$ into the binary image $I_2$, this is noted as
$encode_2(I)$.
\begin{equation}
  \label{continuityEqn}
  L_{color\_continuity} = | encode_2 (I) - encode_2(M(I)) |
\end{equation}

\begin{figure}
    \centering \includegraphics[width=\linewidth]{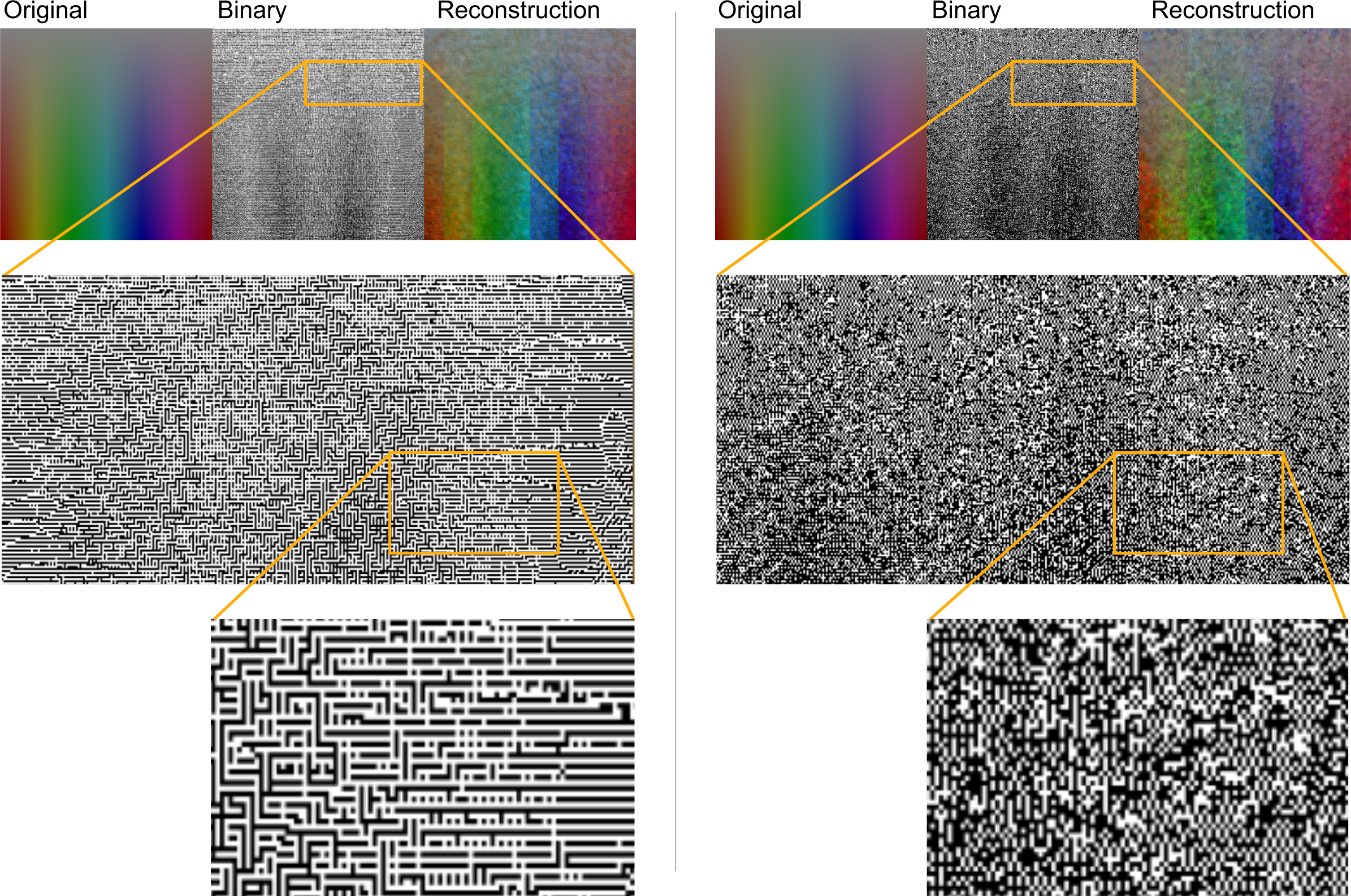}
    \caption{Discretization of the same image using two separately
      trained network to examine color continuity.  %
      When smooth regions of the
      image are examined, we hope to find correspondingly smooth 
      changes in the binarized image. (left): Within the zoomed image,
      notice that the pattern of texture changes rapidly.  (right):
      The change of patterns is far more subtle; this network was trained with
      color continuity constraints. %
    }
    \label{fig:badContinuity}
\end{figure}

In training, in each batch, even numbered images are replaced with a
modification of the previous odd-numbered image.  The simplest
modification operator is used: independent noise per pixel.  This
effectively keep two copies of the image, with and without
perturbation, in the batch --- the goal is ensure that they both have
similar binary encodings.  For example,
Figure~\ref{fig:badContinuity}C shows a poor transformation
where nearby colors are represented with large texture shifts. In (E),
the same region is shown for a network trained with color continuity loss.

\subsubsection{Handling Uncommon Colors}

In training the initial system, real photographs from ImageNet were
used.  However, when the system was tested, often the first test users
performed was on paintings, clip art, or other artwork.  To better
address the distribution of colors  present in these image
classes, we augmented the training set with synthetically generated
images.

The synthetic images were rendered with a background of constant
color.  Two to five shapes (ovals and rectangles) were added to the
image, in random colors, sizes, and aspect-ratios, without considering
overlap.  All color selections were made uniformly randomly from the
set of $2^{24}$ possible choices.  These generated examples were used
to replace $S\%$ of each batch in training.  Numerous settings of
$1\% \le S \le 50\%$ were tested.  Empirically,
$S=10\%$ worked well; it had a noticeable impact on how well
artwork and clip art were binarized while photograph performance was unchanged.

\subsection{Neural Architectures}
\label{others}

Finding the appropriate neural network for this task required
numerous architecture decisions and extensive empirical testing.  For
our study, the decisions can be broadly categorized into three
branches:  convolution sizes, discretization
progression, and network connectivity, specifically the number and size of
hidden layers.

\begin{figure}
    \centering
    \includegraphics[width=\linewidth]{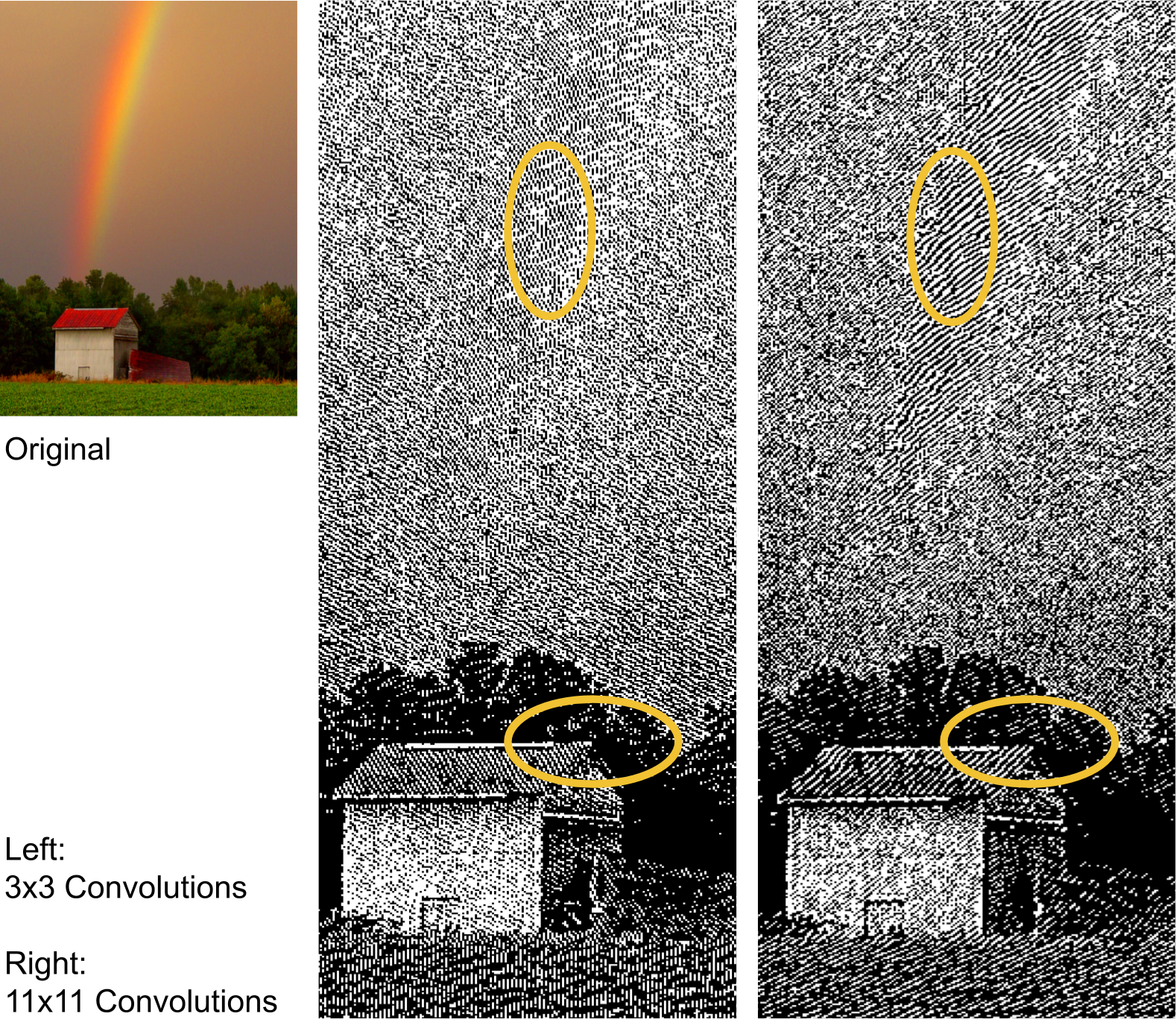}
    \caption{Effects of convolution size on diversity of textures
      produced in the binary image.  In the expanded views, notice
      how the $11\times11$ convolution network is able to yield
      varied textures for the rainbow, tree-lines and roof.  }
    \label{fig:filterSizeComparison}
\end{figure}

Within the image-understanding community, the use of convolutions in
neural networks is standard practice.  The size of the convolution is
vital since it controls the integration of information across the
image %
~\cite{chen2017rethinking,wang2018understanding,NIPS2016_6203}. %
Though image classification
systems commonly use $3\times3$ convolutions, these did not yield the
best results here.  Larger convolutions are able to create a larger
variety of textures, see Figure~\ref{fig:filterSizeComparison}.

Experiments were conducted with convolutions of size $2\times2 -
11\times11$.  Consistently, larger convolution sizes provided more
appealing results than smaller ones.  However, the perceptible
differences between larger sizes diminish.  Because increasing the
size of the convolutions rapidly increases the training times, for our
final system, all the convolutions were $6\times6$.

When discretizing layers in neural networks, often the reduction from
full floating point to a lower bit-depth representation occurs in a
single step. In contrast, as shown in Figure~\ref{fig:firstResults}, the \DDE gradually steps down the discretization
from 8-bpp to 1-bpp.  We also tested the more common single-step
approach, as well as eliminating only selected
intermediate depths.  Though all of the methods provided acceptable
results, the gradual step-down procedure consistently yielded better
results.

Though beyond the scope of this paper, the largest set of experiments
conducted are those that tuned the network architecture and learning
parameters.  Finding ``the right'' number of layers and connectivity
patterns is largely a trial-and-error endeavor.  For example, we
experimented with using (2-20 layers), as well as a wide variety of
number of channels (10-150) per layer. The majority of early
exploration also employed equal size pre-encoder and decoders, as is
common practice.  An unexpected
finding was that a large decoder was not only unnecessary, but could
be detrimental.  In the final system, the decoder and pre-encoder were
asymmetric.  The decoder was reduced to only 2 layers, while the
pre-encoder remained 10 layers.  We postulate that a smaller decoder
is beneficial because it forces the binary image to appear as close as
possible to the final image by not permitting large transformations of
the binary image to create the reconstruction. Full details on the
selected network architecture are provided next.

\subsection{Learning Summary}
\label{summaryTrain}

In designing the final system, numerous, diverse architectures, loss
functions, and network training meta-parameters (gradient descent algorithm,
non-linear activation functions, learning rate, \emph{etc.})  were
explored.  In total, 980 different networks were trained.  The single
network chosen is specified in
Table~\ref{architecture2}.  The final loss to be minimized was ($\alpha=0.1$ and $\beta=0.1$):

\begin{equation}
  \begin{split}
   Total\_Error & =  L_2(I_{RGB}-I'_{RGB}) \\&+ (\alpha \times L_{relative\_intensity}) \\& + (\beta \times L_{color\_continuity})
  \end{split}
\end{equation}
\vspace{0.1in}

For our training set, 90\% of the examples were photographs from
ImageNet. The remaining 10\% were synthetically generated.  The 
same Tensorflow parameters for training described earlier were used
again.  The training time on a single
Nvidia-P100 GPU was approximately 36 hours.

For the final network selection, we whittled the trials from the GANs,
style transfer networks, and the 980 architectural variants to
25. This pruning was based on a combination of a cursory scan of the
resulting images and the magnitude of the reconstruction errors.  With
these 25 sets of remaining images, an independent user-interface
expert was asked to examine the images generated by all 25 networks
and select those that created the binary image that best represented
the original.  Neither the reconstructed image nor any of the other
network feature were considered or even revealed.  The network that
obtained the most votes is shown in Table~\ref{architecture2}.  A
total of 150 images were evaluated in this phase.  During the
evaluation process, the majority of the initial 25 networks received
no votes, suggesting that a few networks clearly outperformed the
rest\footnote{
Having conducted this extensive network exploration, we must re-emphasize that
training neural networks is a fundamentally stochastic procedure.  The
performance of the same network can vary based on the order the
training samples are presented, the training times, the initial
randomly selected network weights, and the meta-parameters, to name a
few.  %
This fact, coupled with the qualitative nature and
inherent noise in ascertaining which binarized image appears better
than another, makes it difficult to definitively assert that the single
architecture selected is better than all the rest.  Due to stochasticity in training, a different set of
parameters or random initial weights might have changed the
performance.  If replicating the system, we suggest 
training several networks, even with the architecture and parameters
found here, to overcome outliers in performance.
}.

\begin{table}
 \renewcommand{\arraystretch}{1.2}  
 \caption{Final architecture. Each layer uses \underline{($6\times6$)} convolutions.\\Note the asymmetric pre-encoder and decoder sizes.\\The final binarized output is marked with *.\\(Difference from previous architecture is highlighted.)
 }
  \label{architecture2} 
  \footnotesize
  \begin{tabular}{c|ccccc}
    & \rotatebox{0.00001}{input} 
    & \rotatebox{0.00001}{\makecell[l]{layers, \\channels\\per layer}}
    & \rotatebox{0.00001}{\makecell[l]{activation\\function}}
    & \rotatebox{0.00001}{\makecell[l]{activation\\ function\\ in last layer}}
    & \rotatebox{0.00001}{output}\\
    \hline
    \hline    
    \makecell{Pre\\Encoder}         & \makecell[c]{24 bit\\\NXN}  & 10, 128 & relu & relu &\makecell[c]{128 float\\\NXN} \\
    \hline    
    \makecell[c]{Down\\Discretization\\Encoder}  & \makecell[c]{128\\float\\\NXN}    & 8, 1 & \makecell[c]{tanh\\discrete$_{256}$-\\discrete$_{4}$} & \makecell[c]{tanh\\discrete$_{2}$} &
                         \textbf{\makecell[c]{1 bit\\\NXN(*)}}\\
    \hline    
    Decoder             & \makecell[c]{1 bit\\\NXN}     & \cellcolor{yellow}2, 128 & relu & tanh &  \makecell[c]{24 bit\\\NXN}\\
  \end{tabular}
\end{table}

\begin{figure*}
  \centering
 \includegraphics[width=\textwidth]{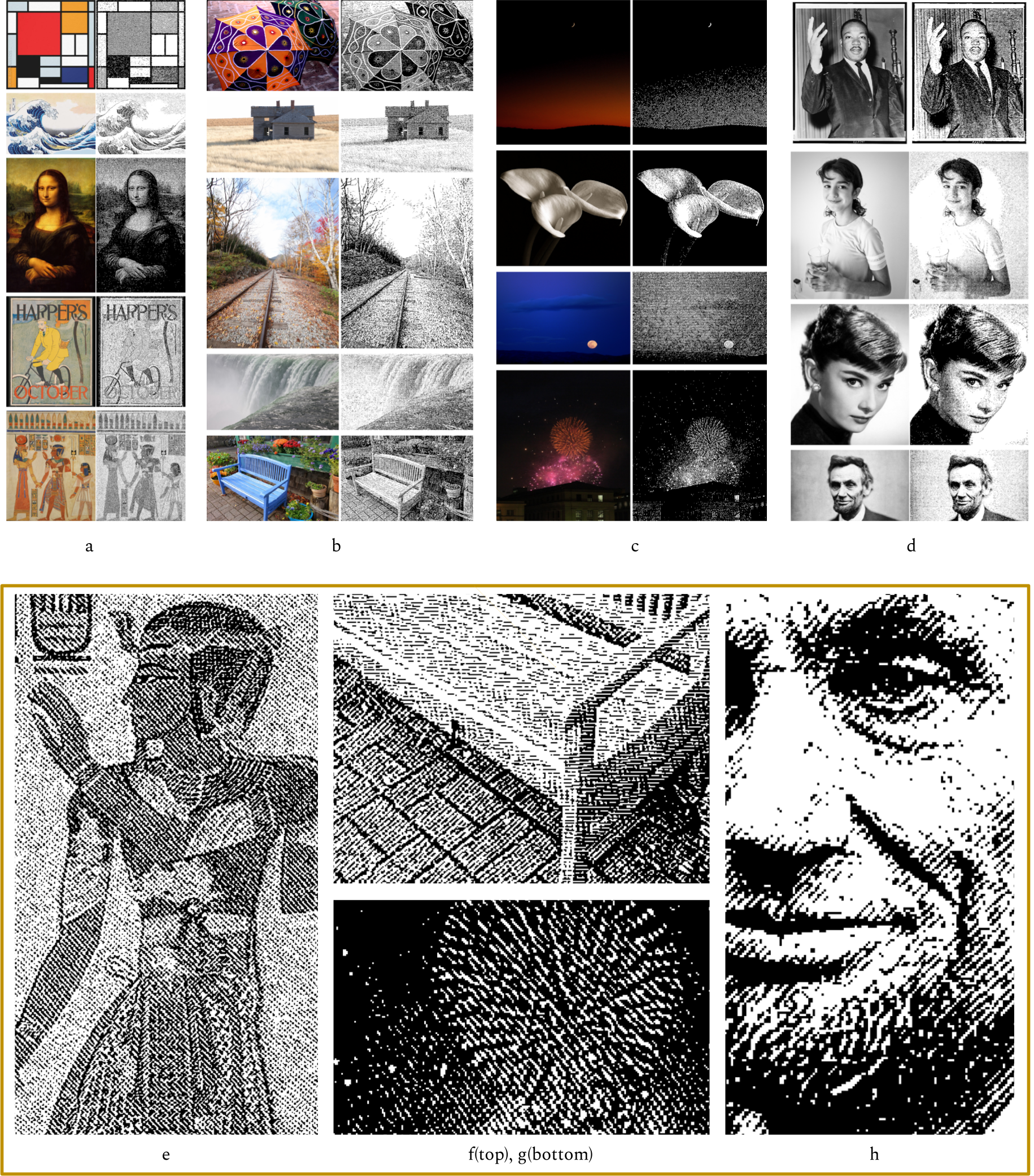}

  \caption{Results in four categories: (a) paintings and art, (b)
    photographs, (c) dark images, and (d) black and white source
    images. Enlarged views of the bottom-most images in each category
    shown within the orange rectangle (e-h).  Please see text for full description. }
  \label{fig:trialHResults}
\end{figure*}

\begin{figure*}
  \centering
  \footnotesize
  \begin{tabular}{cc}
    
    \includegraphics[height=1.0in]{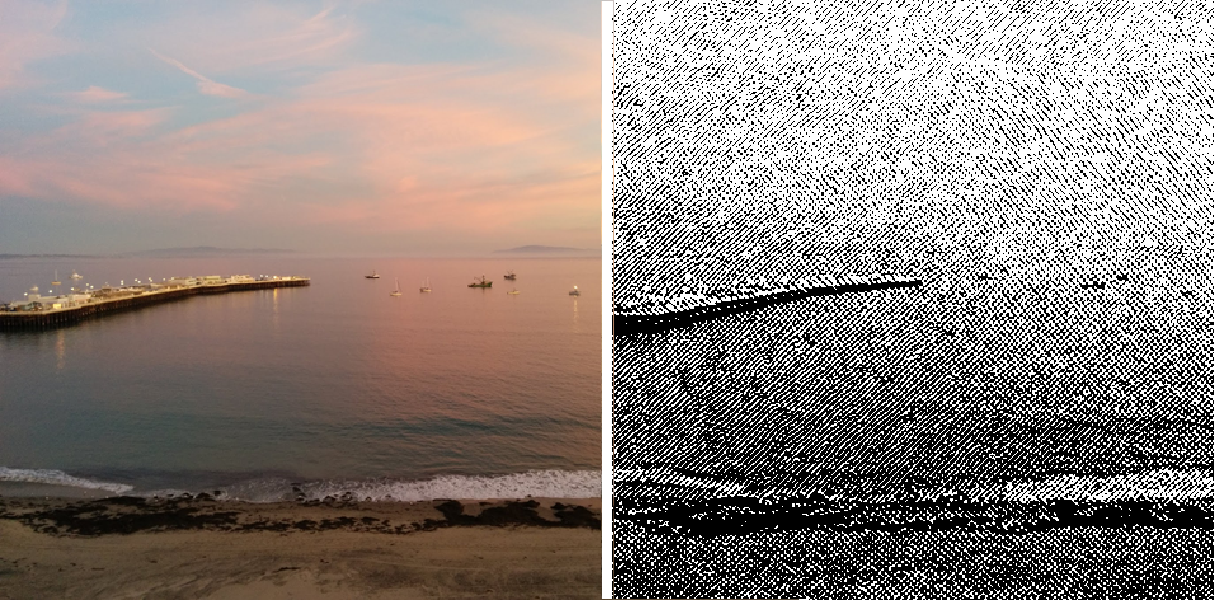}&
    \includegraphics[height=1.0in]{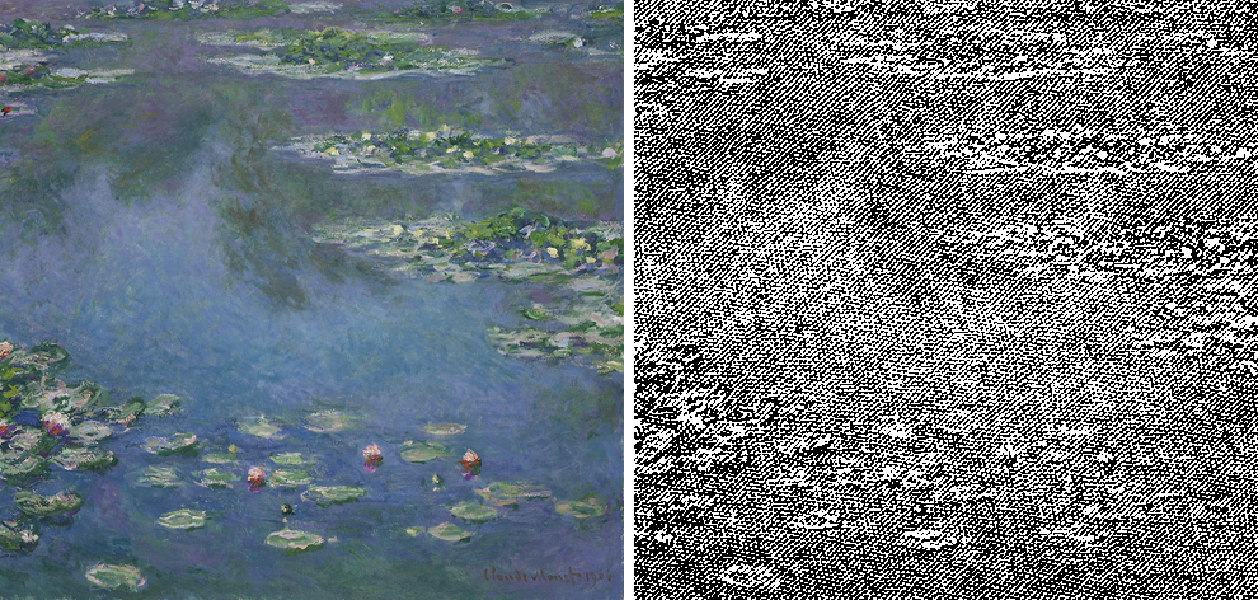}\\
    a & b\\
    \includegraphics[height=1.0in]{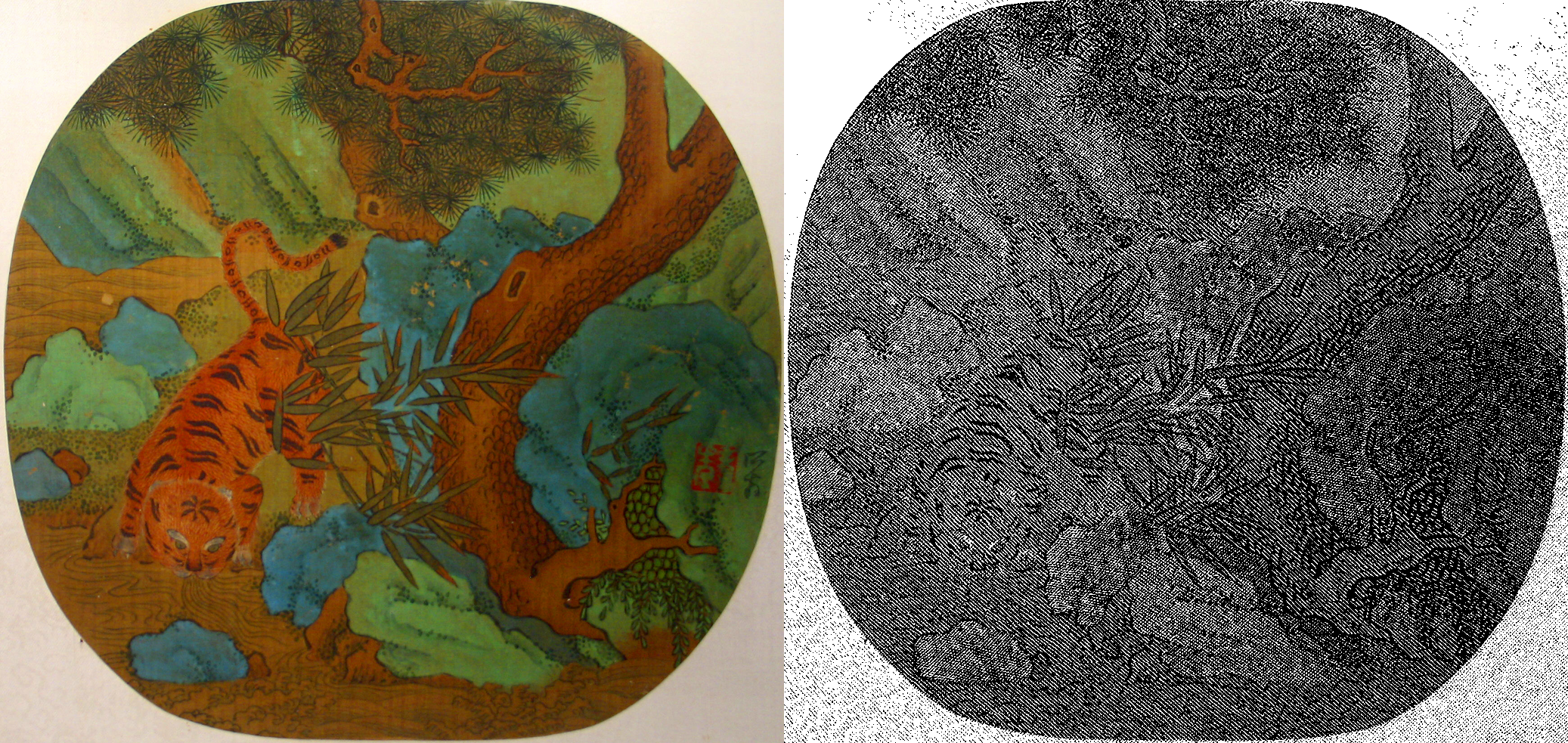}&
    \includegraphics[height=1.0in]{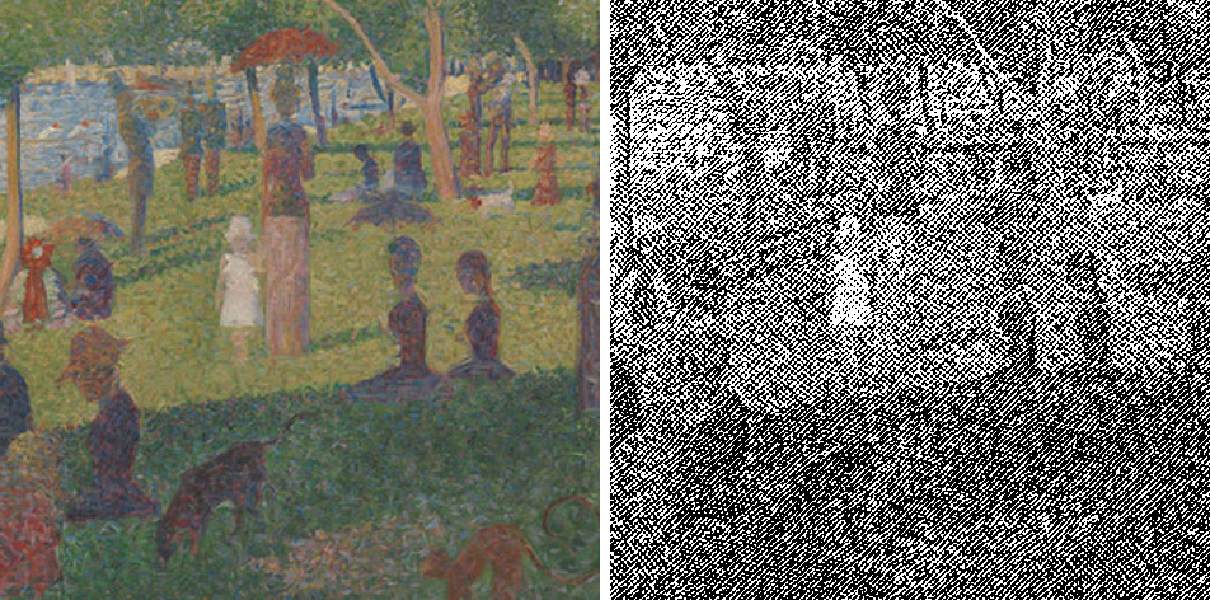}\\
    c&d\\
    \hline
    ~\\
    \includegraphics[height=1.0in]{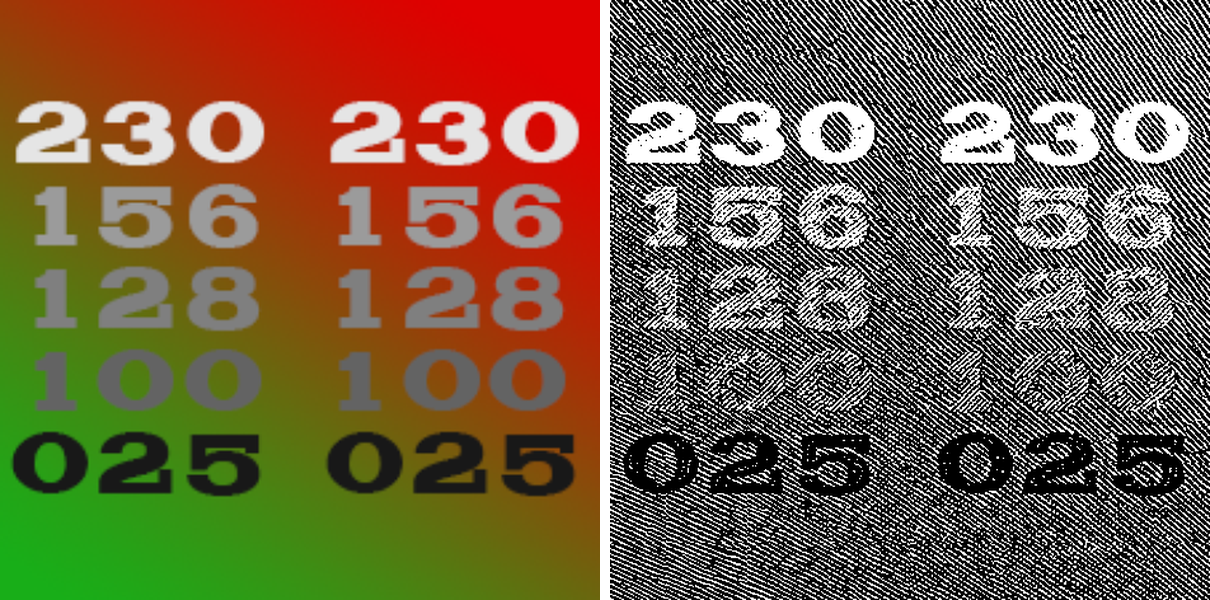}&
    \includegraphics[height=1.0in]{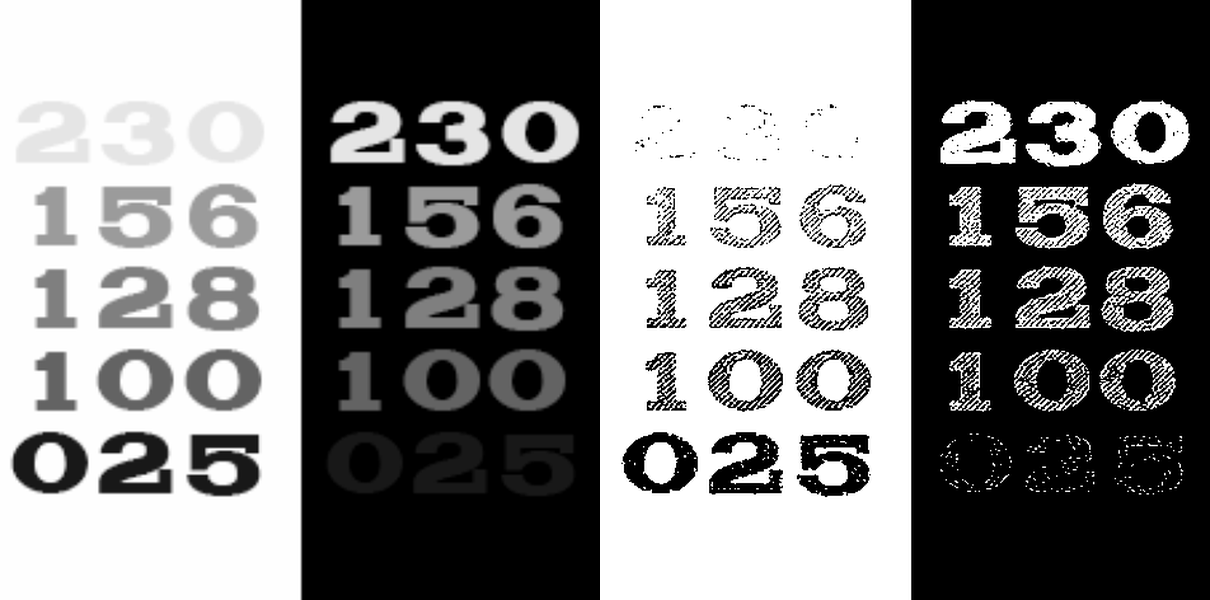}\\
    e & f\\
    \includegraphics[height=1.0in]{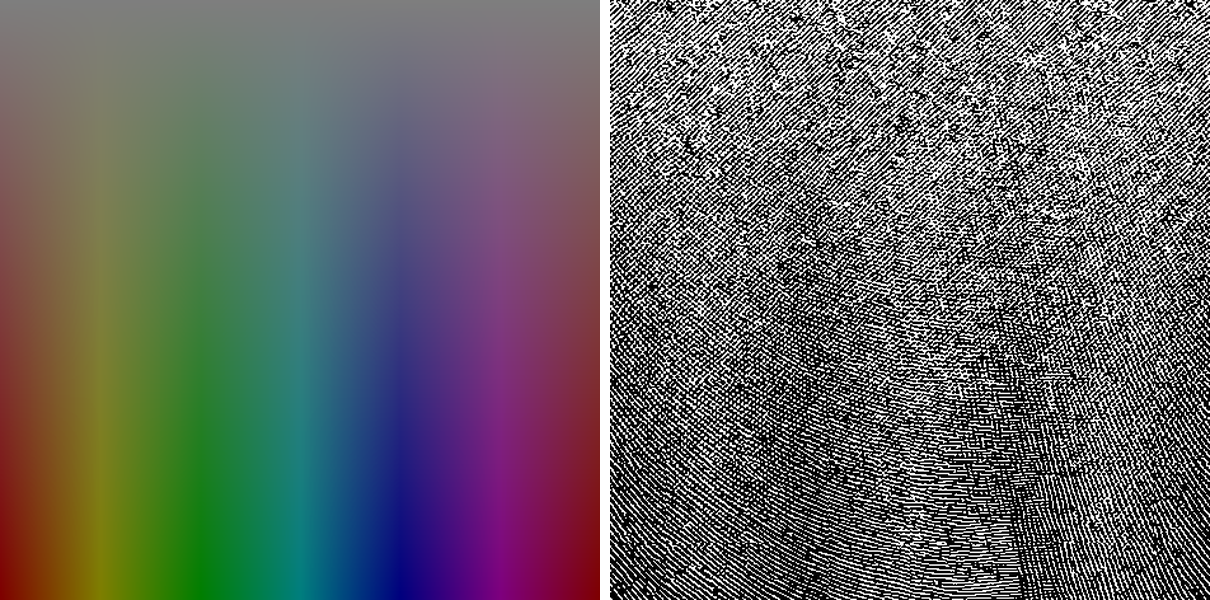}&    
    \includegraphics[height=1.0in]{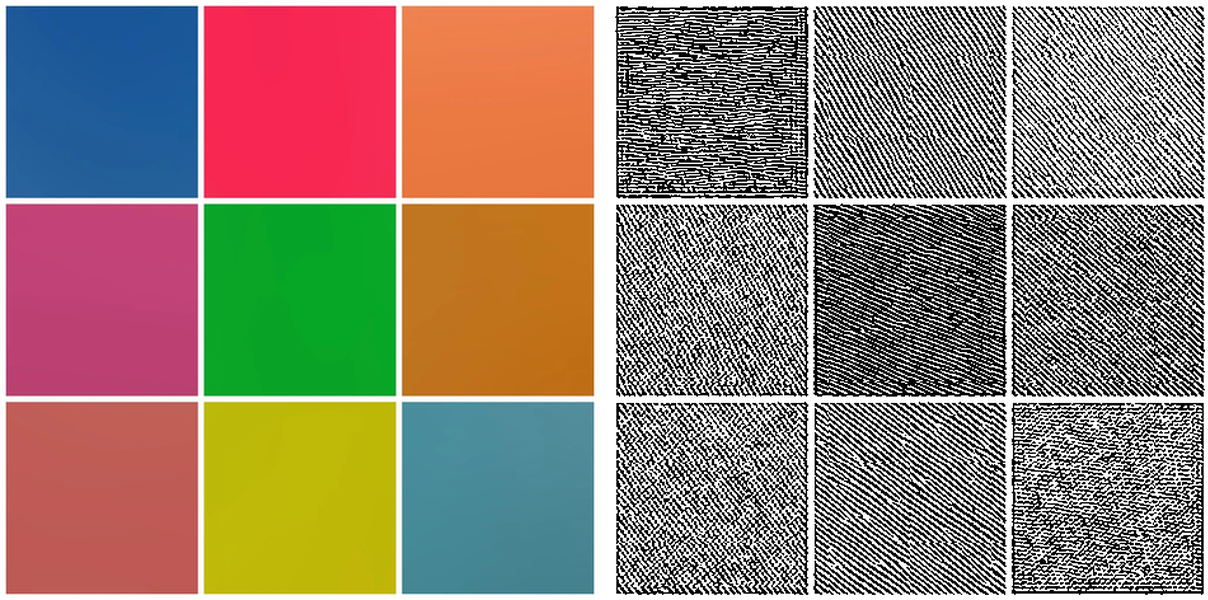}\\
    g&h\\ 
  \end{tabular}
  \caption{Limitations of our approach.  a,b:
    A lack of contrast yields too little sky/water distinction in the photograph and in Monet's \emph{Water Lilies} (1914). c:
    \emph{Tiger Drinking from a Stream} (Qing Dynasty period) the tiger
    appears camouflaged due to the similar texture representation of
    these shades of orange and brown.  d: Seurat's pointillism; the
    colors in close proximity to each other result in no coherent
    patterns.  Bottom section: four explicit tests to examine
    smoothness and color representation range.  e,f: Various backgrounds with foreground
    numbers in the gray-level indicated by the number.  g,h: Smooth
    and abrupt color boundaries.}
  \label{fig:extraTests}  
\end{figure*}

\section{Experiments and Results}

Figure~\ref{fig:trialHResults} shows the results of using the selected
final network (Table~\ref{architecture2}) on 18 images drawn from a variety of sources.  (a) shows
the results on artwork that contains colors outside the range of
standard photographs.  In the photographs in column (b), note that the
changes in hue as well as intensity are successfully captured in the
binary representation.  With the darker images in (c), the intensity
changes are vital to capture.  Note that the ``pop-out'' brightness of the
moon on the night-sky is captured (photographs 1 \& 3 in column c).  The black
and white originals in (d) demonstrate that although the network was
not trained on any grayscale images, the binarization is successful.

In Figure~\ref{fig:trialHResults}e, expanded views show details of the
textures synthesized for four results (the bottom image of each
column).  In (e), notice the variety of textures and clear boundaries
between distinct regions.  In (f), similar textures are employed to
represent all the wood pieces in the bench; these give the sense of
similarity between the pieces and the continuity of each slat.  In
(g), the different fireworks are easily distinguishable through the
use of textures.  In (h), the same (or similar) textures are
automatically chosen for the beard, resulting in natural appearing
shading.  Additional results are shown in Figure~\ref{teaser} and the
Appendix.

In Figure~\ref{fig:extraTests}(top), we provide four examples to
elucidate the limitations of our approach.  In (a), the low contrast
in the image obscures the water/sky boundary in the binarized image.
In (b), though the regions are identifiable, without color, the image
is difficult to interpret.  In (c) the tiger is difficult to see due
to similarities in the orange and brown textures.  In (d) we see the
most insidious of cases for our system: pointillism. In Seurat's
\emph{A Sunday on La Grande Jatte} (1884), the painting is created
with tiny dots of various hues and saturations. The rapid changes in
color as well as the many discontinuities are difficult.  Reducing the
size of the image (and thereby averaging the pixels) or simply
blurring the image before generating the binary representation will
improve results.

Finally, we present four images specifically created to test our
system, Figure~\ref{fig:extraTests}(bottom). These provided controlled tests
to measure the diversity of colors being represented and how color
transitions are handled.  Figure~\ref{fig:extraTests}e is a gradient
background image with text written in various intensities (the number
is the intensity value out of 255) (f) repeats this test with a pure
black and white background.  Figure~\ref{fig:extraTests}(g\&h)
examine smooth and abrupt color changes.

\section{Conclusions and Future Work}
\label{conclusions}

We presented an automated method to replace a color or grayscale image
with a set of binary textures that not only represent the original
image's colors and patterns well, but adhere to the aesthetic and
pragmatic desiderata stated in Section~\ref{related}:

\begin{enumerate}
    \item \emph{Different perceived colors have different texture
    representations}.  This is satisfied through the training
      procedure; the system is trained to ensure that we can
      recreate the original color image from the binary
      representation.
    \item \emph{Colors that are visually similar are represented by
    similar textures.}  This was explicitly addressed with the
      continuity loss described in Section~\ref{continuity}.
    \item \emph{Different images have the same color represented
    similarly}.  This can be seen by the ability to reconstruct the
      original's colors using the same ``decoder'' module.
    \item \emph{The textures introduced to represent colors are not visually
      distracting}.  See Figure~\ref{fig:trialHResults} and the Appendix.
    \item \emph{The approach should be extendable to variable
    bit-rates.}    The extension to other bpp rates is built into the system and is trivial to instantiate.
      For example, to reduce the final representation to 4-bpp instead
      of 1-bpp, we need only remove the last 3 layers in the \DDE and
      retrain.  
    \item \emph{The proposed method should work equally well for
    grayscale images as for color images.}  As shown in
      Figure~\ref{fig:trialHResults}, grayscale images are treated the
      same as color images; no extra steps were taken to handle these.
    \item \emph{No manual specification of good vs. bad binarizations}.
      The system is entirely
      \textbf{self-supervised}.  There were no manually created
      examples of good and bad binarizations.
      Instead, the system was trained through the principle of information
      preservation.  
\end{enumerate}

\begin{figure}
    \centering
    \footnotesize
    \begin{tabular}{>{\centering\arraybackslash} m{0.6in} >{\centering\arraybackslash} m{1.7in} >{\centering\arraybackslash} m{.65in}}
      \shortstack{\underline{small decoder}\\ \: \\pixel error\\=17.3\\\:}&
      \includegraphics[width=1.65in]{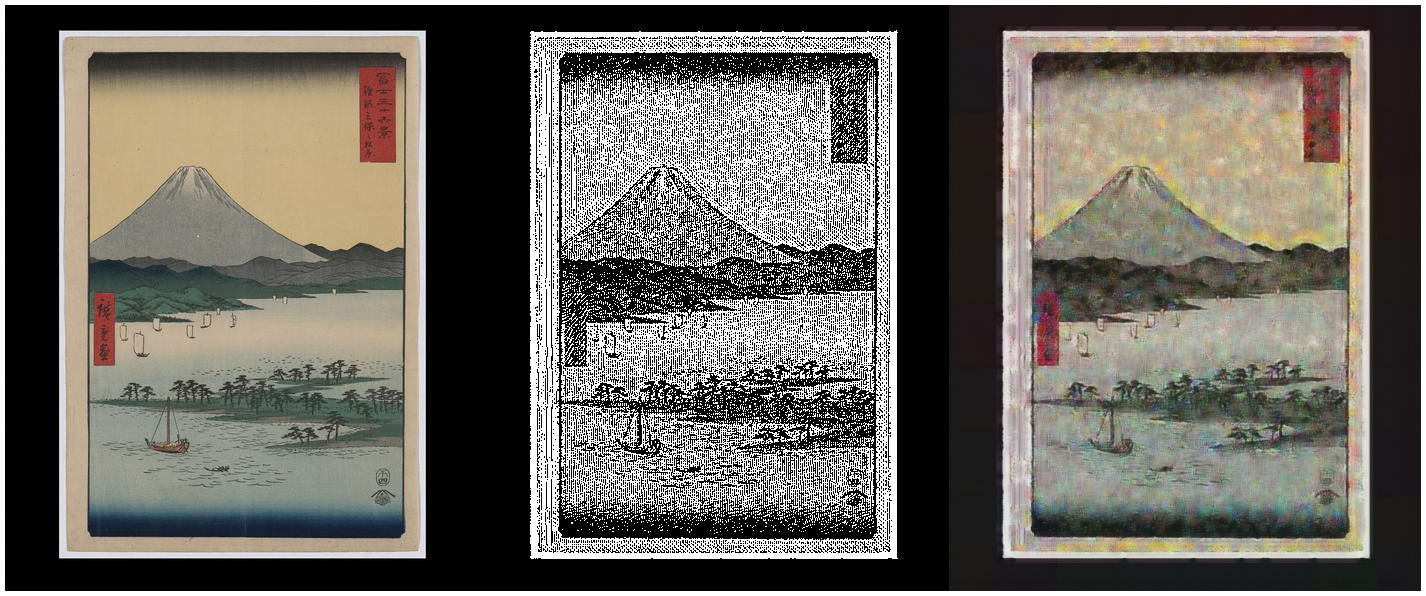}&
      \includegraphics[width=.55in]{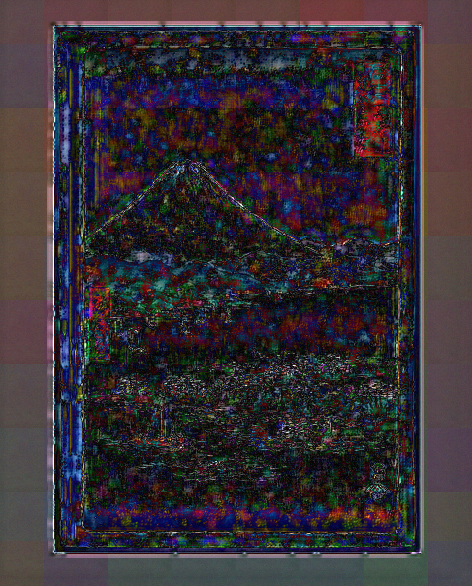}\\
      \shortstack{\underline{large decoder}\\ \: \\pixel error\\=8.3\\\:}&      
      \includegraphics[width=1.65in,height=.7in]{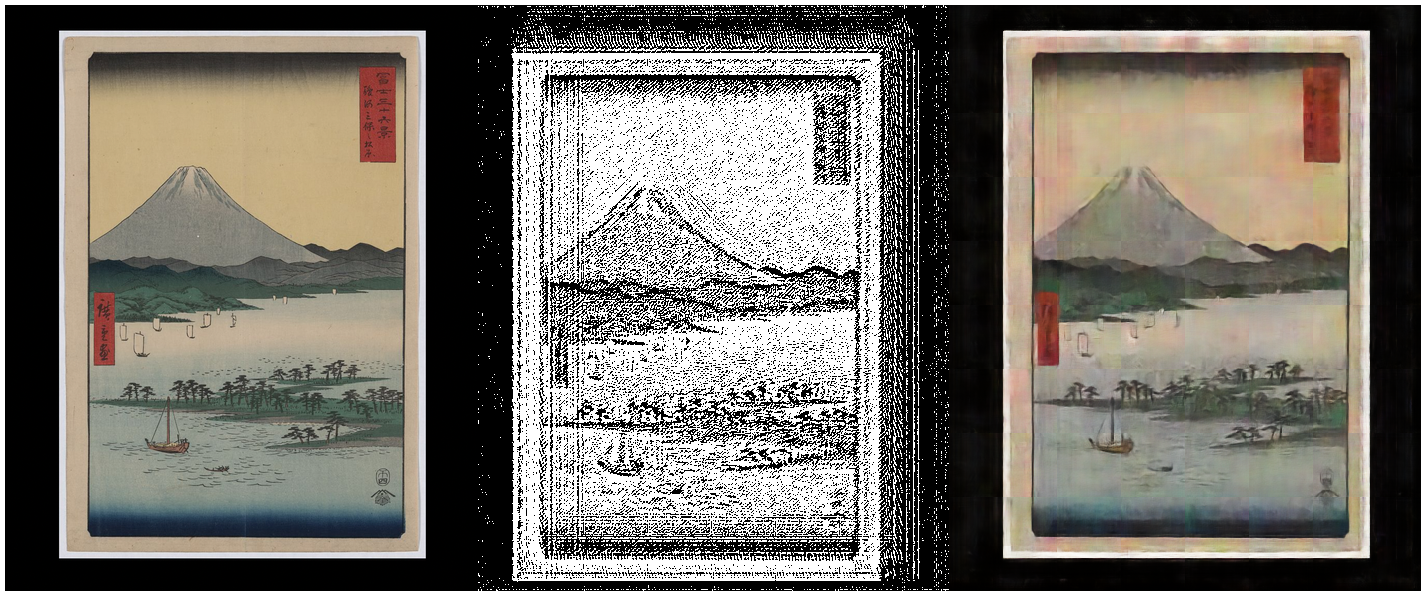}&
      \includegraphics[width=.55in]{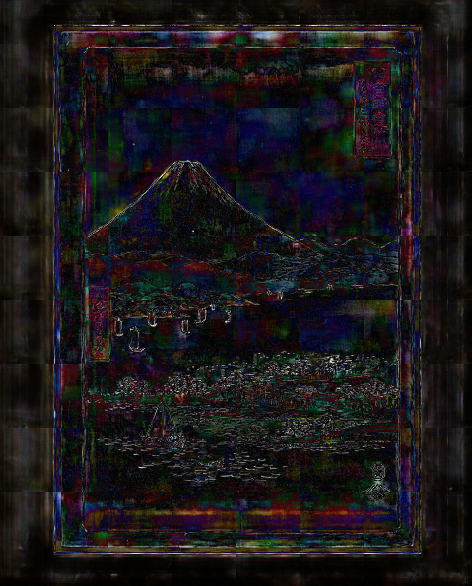}\\     
      \:&\\      
      \hline
      \:&\\
      \shortstack{\underline{small decoder}\\ \: \\pixel error\\=21.2\\\:}&      
      \includegraphics[width=1.65in]{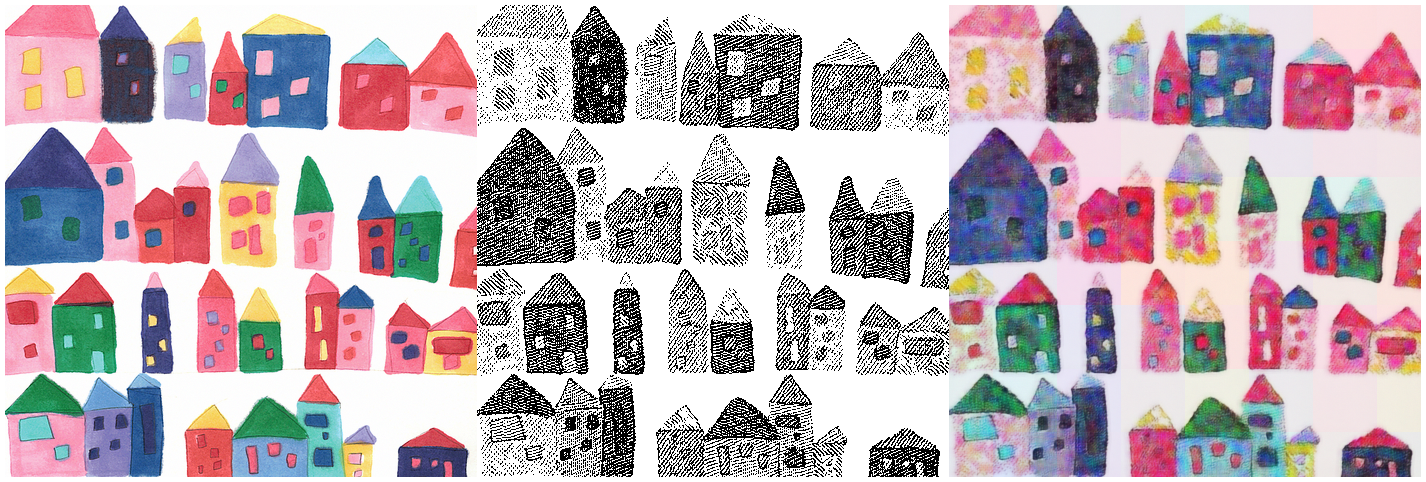}&
      \includegraphics[width=.55in]{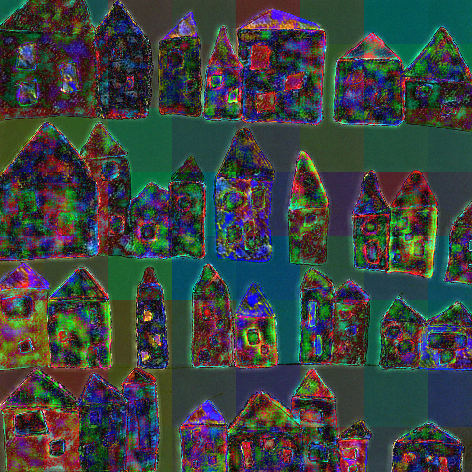}\\     
      \shortstack{\underline{large decoder}\\ \: \\pixel error\\=11.7\\\:}&            
      \includegraphics[width=1.65in]{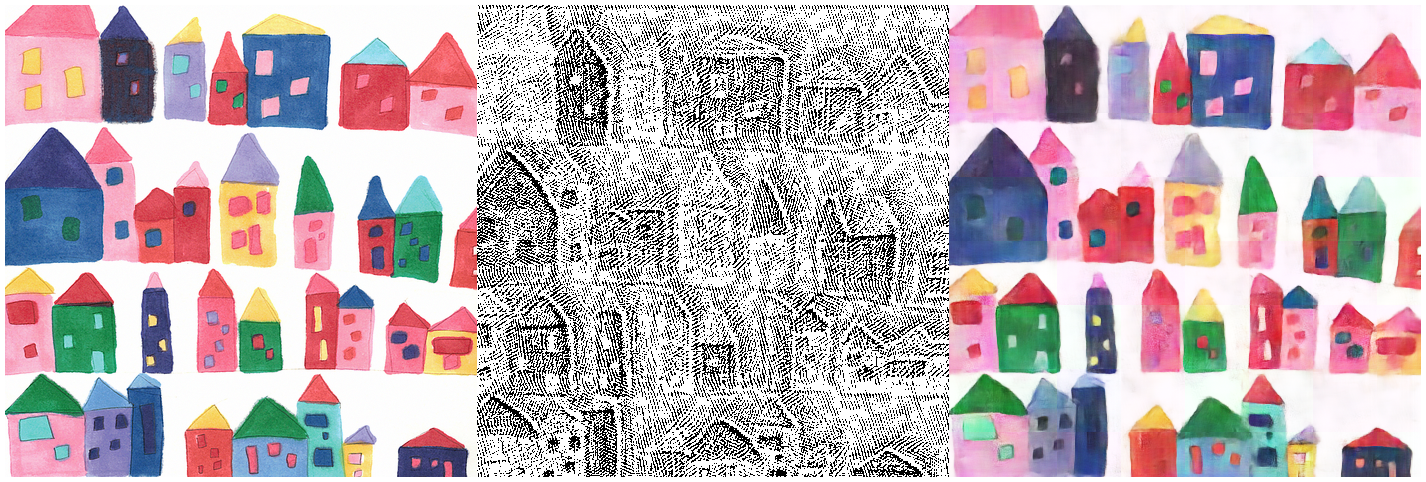}&
      \includegraphics[width=.55in]{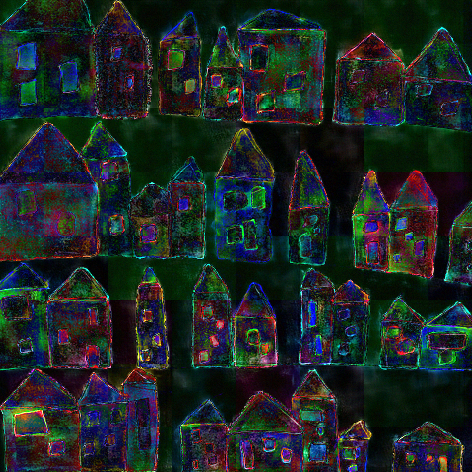}\\
  \end{tabular}
  \caption{ Notice the improved reconstruction with a large decoder
    network (10 layers) vs. the small decoder (2 layers).  The average pixel error (per channel) of the
    reconstructed image drops with the larger decoder.  In each
    triplet: (left) original image, (middle) binary image, (right)
    reconstruction after the decoder.  In the last column, a 3x
    brightened difference-image between the original image and its
    reconstruction.  The improved reconstruction can degrade the
    faithfulness of the binarized image to the original, since with a
    larger decoder, the binary image can be more transformed.  Top:
    wood-print \emph{Suruga miho no matsubara} by Ando Hiroshige, 1858.
    Bottom: watercolor, children's artwork.  }
    \label{fig:recons}
\end{figure}

Though an integral portion of our system, the reconstruction error was
solely employed during training. In usage, we discard the decoder and
no longer compute the reconstruction error.  Nonetheless, in some
applications, the reconstruction may be a valuable product as well,
\emph{e.g.} for low-bandwidth transmission of an image that can be
converted back into color.  In this regard, we note that there are
explicit knobs for controlling the quality of the reconstruction
vs. the quality of the binarization.  The most effective is the size
of the decoder (which was reduced in the final architecture to only two
layers from the initial tests with 10 layers).  The more
transformations the binary image is permitted before reconstruction,
the better the reconstruction, but the less the binary image is
constrained to represent the original image.  See
Figure~\ref{fig:recons}.  This is an open topic for further research
and a new set of applications.

When we allowed the original images to be pre-processed with simple
operations (contrast, brightness, saturation, etc), improved results
were consistently possible.  However, for this paper, we applied
\emph{no pre-processing} on the original images to keep the
experiments controlled.  Methods to automatically pre-process the
images before they are given to the neural network is an open, and
straightforward, area for future exploration.

Looking towards the aesthetic and artistic potential of this system,
consider the image of the blue bench in Figure~\ref{fig:trialHResults}. The
strokes appear reminiscent of wood-cuts or the hatching techniques
often used in illustrations within historic books and papers (see
Appendix).  We have not attempted to limit the family of
textures permitted in the binarized result, for example, to allow only
textures that appear as hatches or cross-hatches.  Exploring methods
to constrain the set of allowed textures, either programatically or
manually, may yield a fruitful avenue for artistic exploration.

\bibliographystyle{ACM-Reference-Format}
\bibliography{arts}

\appendix
\onecolumn
\section{Appendix}

The next 9 pages illustrate the binarization of artworks and
photographs.  They are shown expanded so that the textures created can
be examined.  Please note how different colors are handled. The sharp
boundaries in the original image are recreated through the use of
distinct textures.  Gradual fades are represented by textures that
change slowly, thereby allowing for seamless overlap.   
~\\
~\\
~\\
\begin{figure*}[h]
  \centering
   \includegraphics[width=\textwidth]{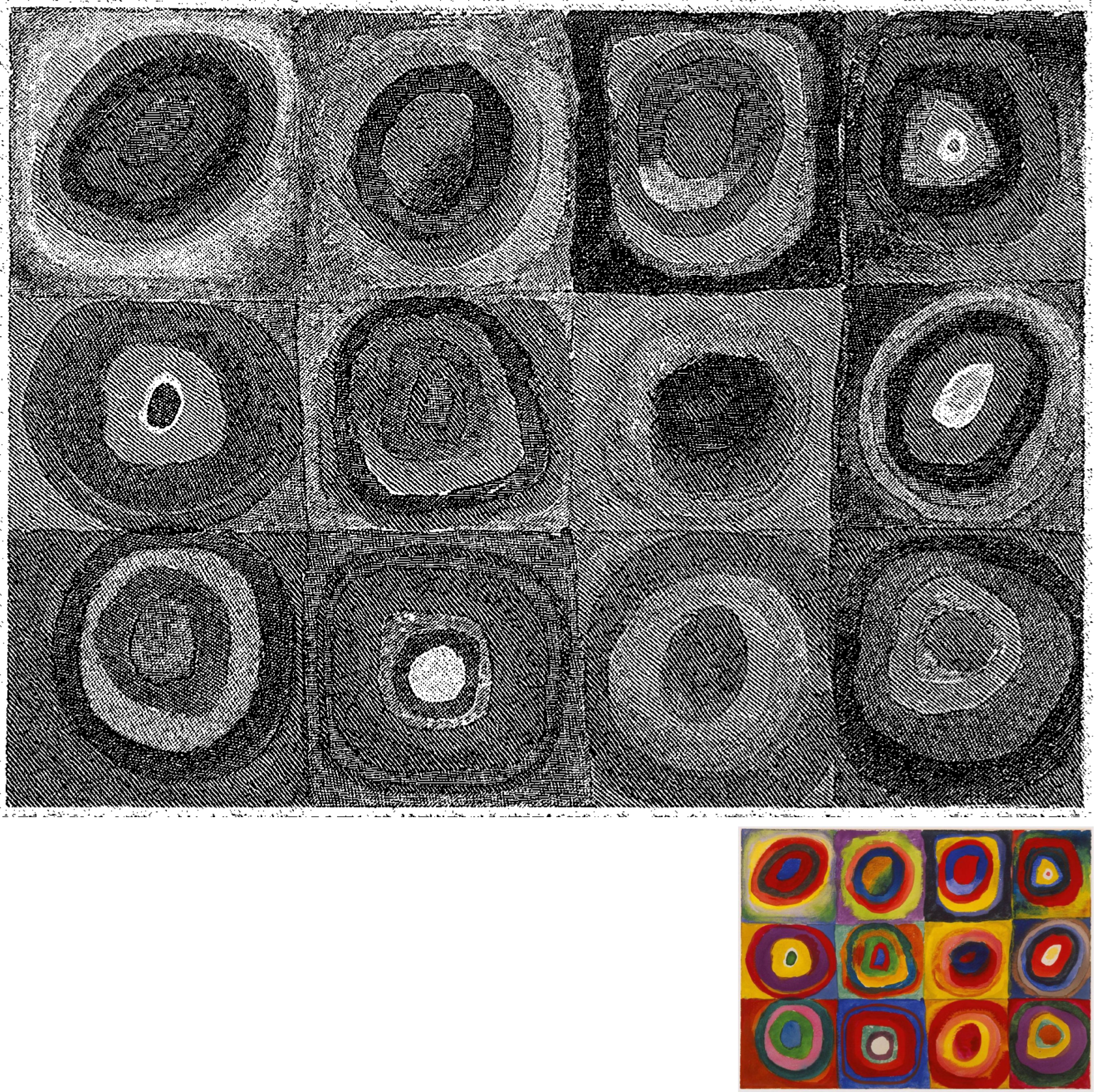}
  \caption{\emph{Color Study. Squares with Concentric Circles} (1913)
    Wassily Kandinksy.}
\end{figure*}
\clearpage 

\begin{figure*}
  \centering
   \includegraphics[height=8in]{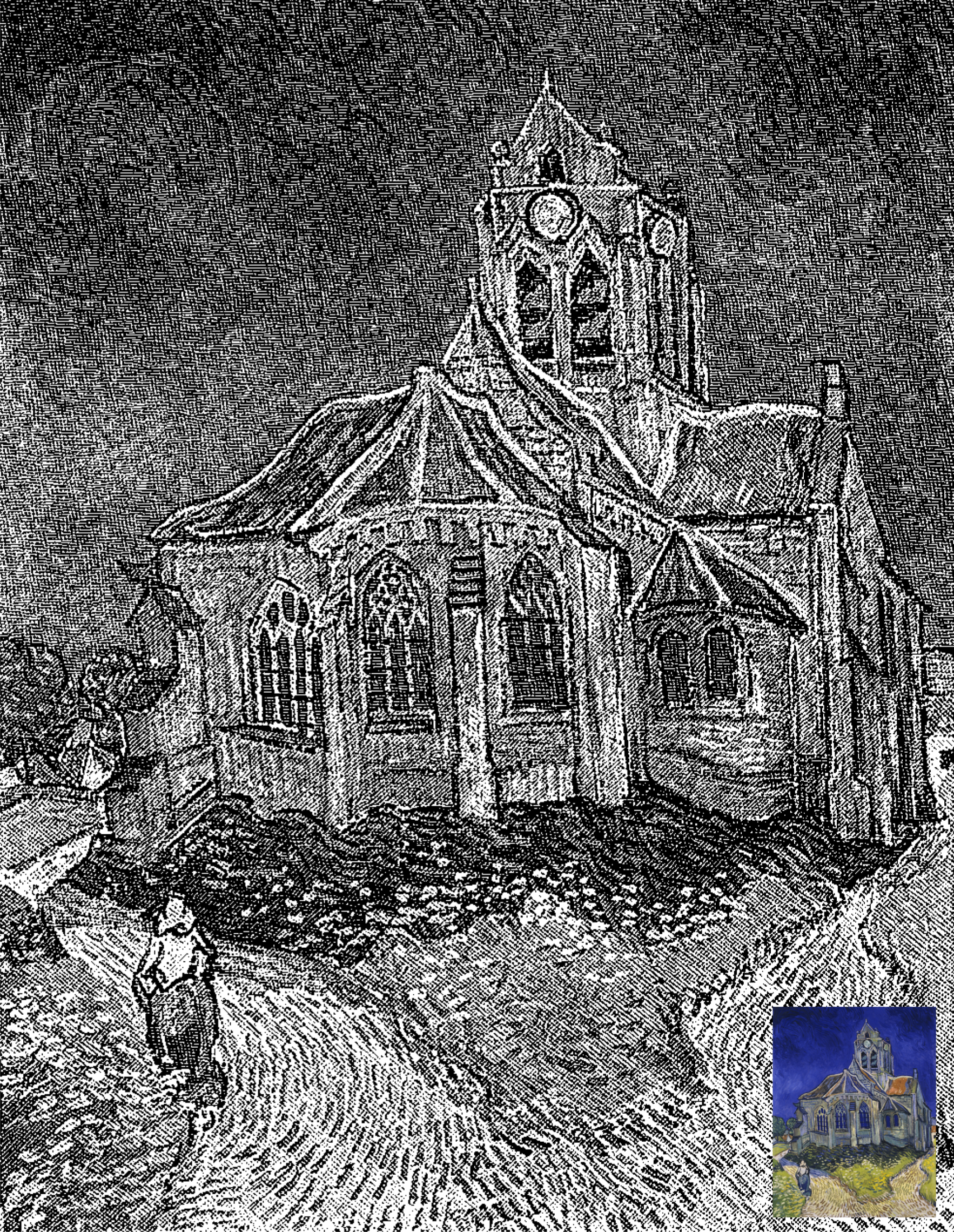}
  \caption{\emph{The Church at Auvers}
    (1890) by Vincent van Gogh}
\end{figure*}

\clearpage 

\begin{figure*}
  \centering
   \includegraphics[width=\linewidth]{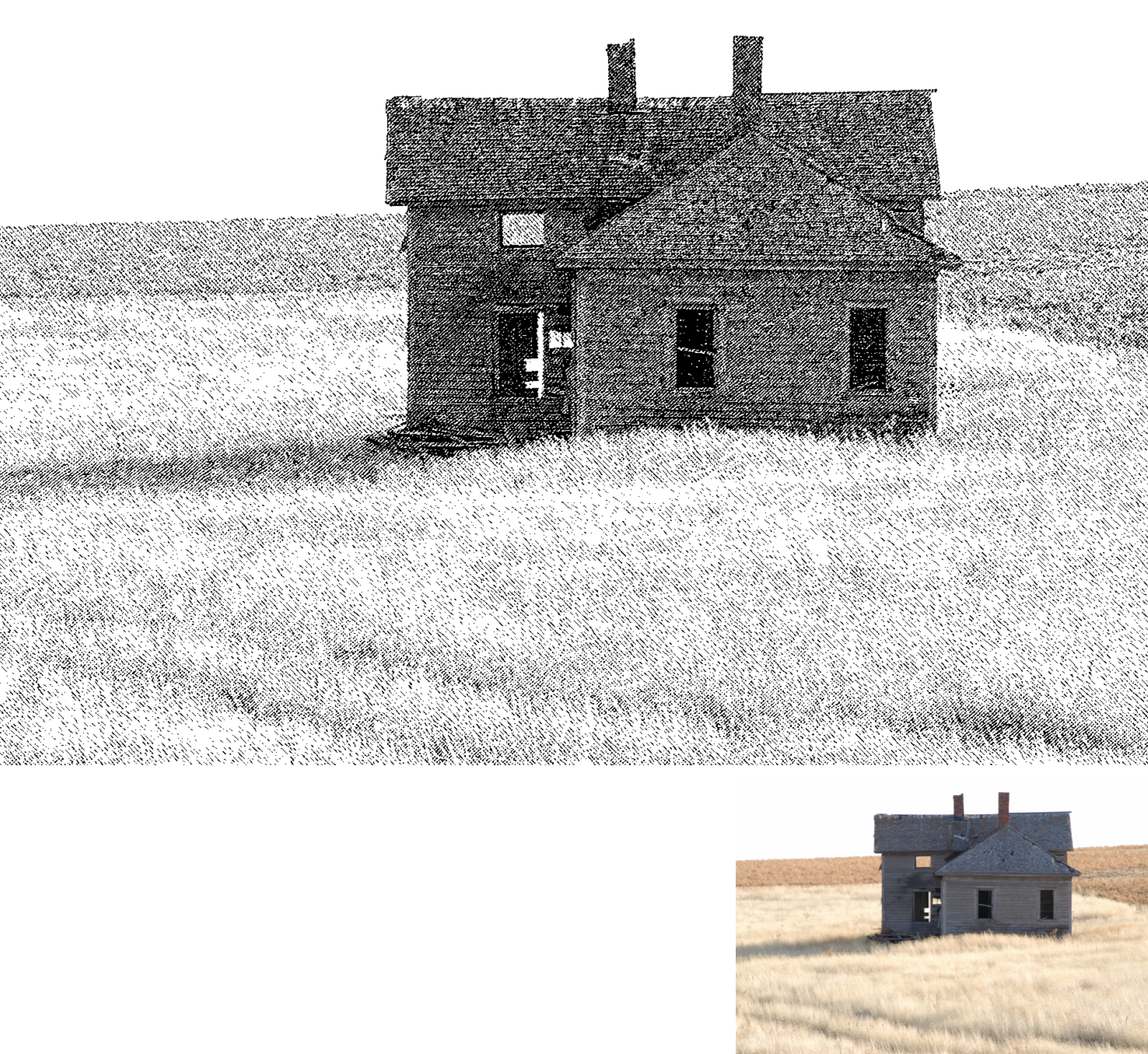}
  \caption{Photograph of a burned house (2003)}
\end{figure*}
\clearpage 

\begin{figure*}
  \centering
   \includegraphics[height=8in]{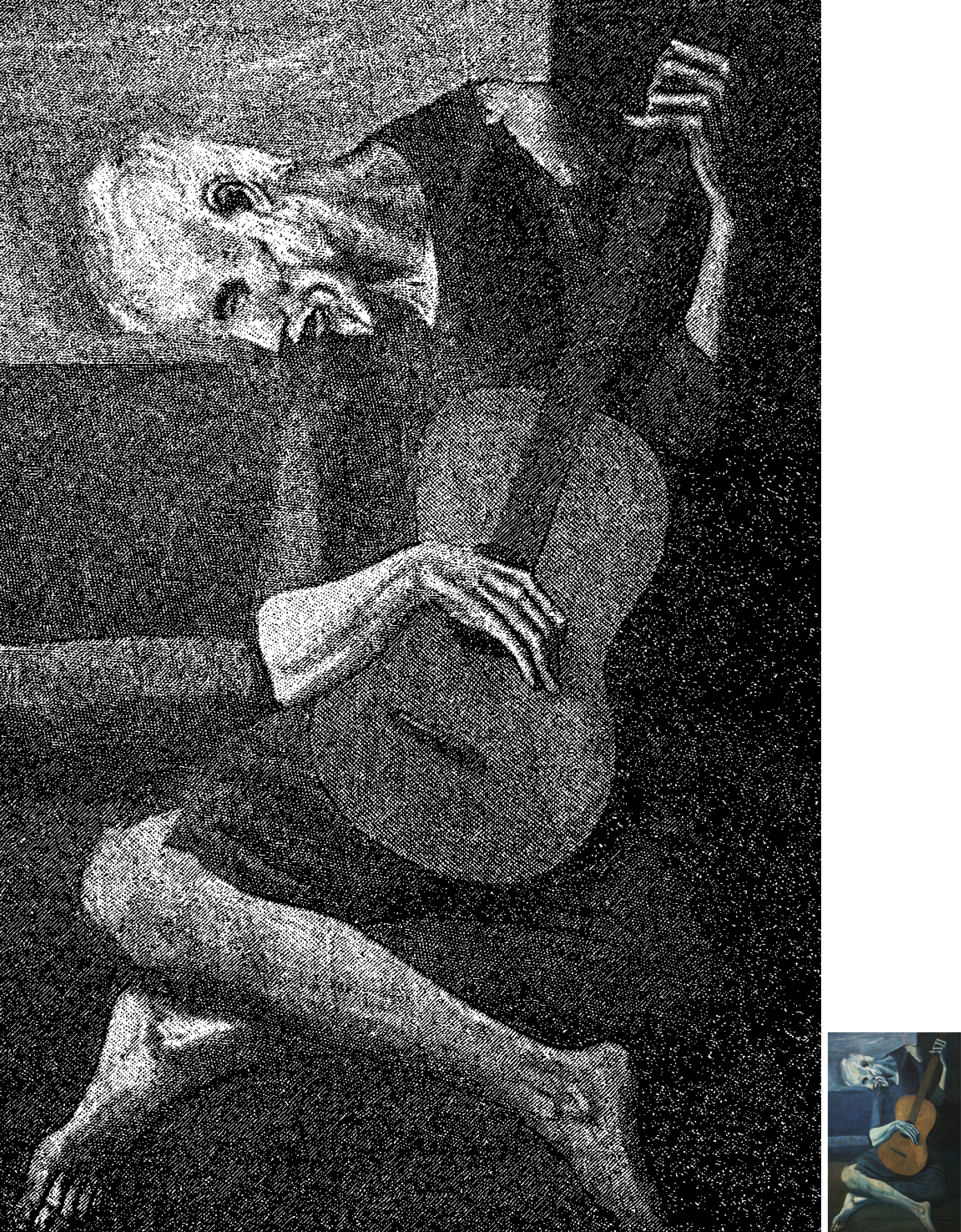}
  \caption{\emph{The Old Guitarist} (1903) Pablo Picasso}
\end{figure*}
\clearpage 

\begin{figure*}
  \centering
   \includegraphics[width=\linewidth]{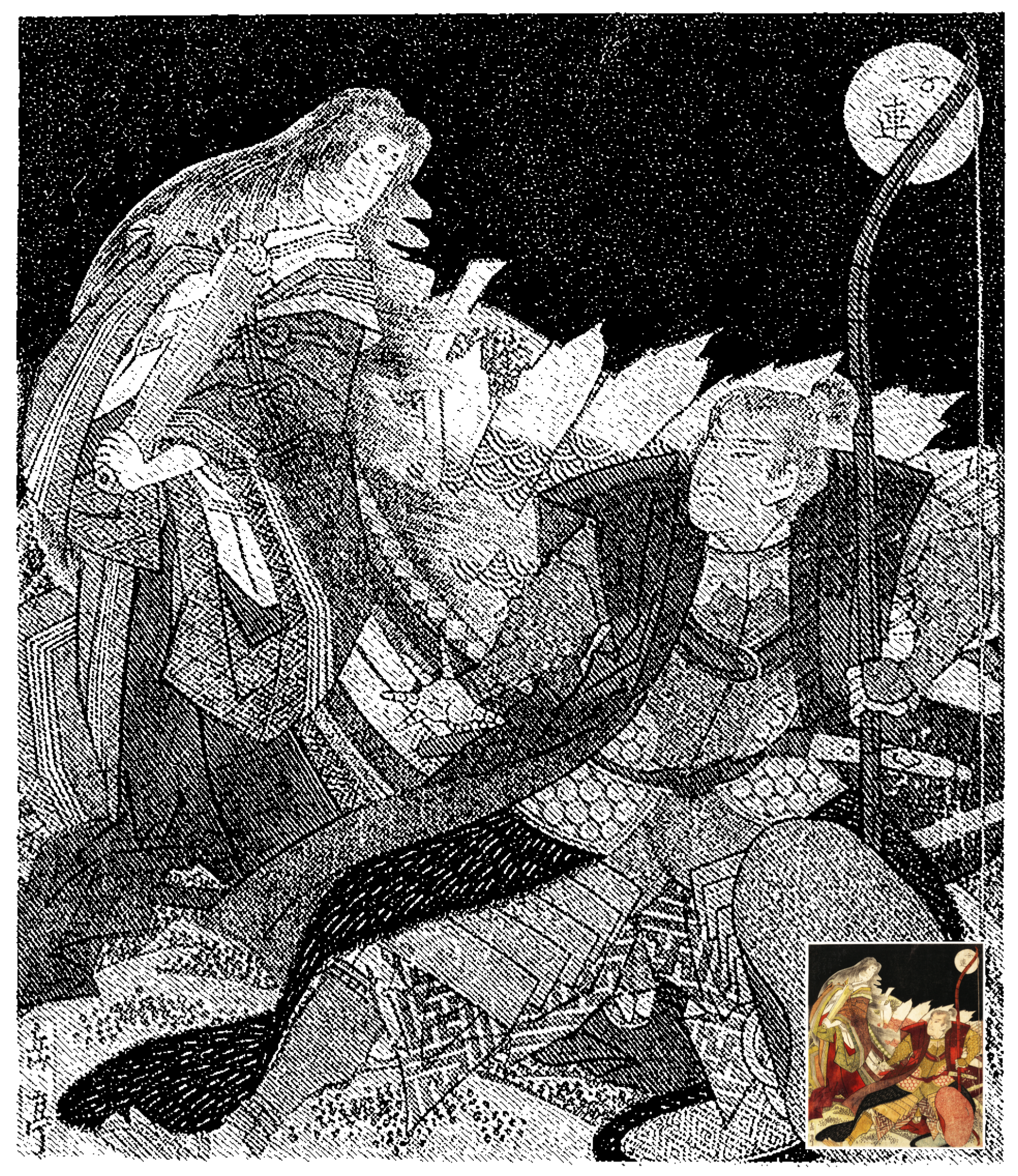}
  \caption{\emph{The Warrior Miura-no-suke
    Confronting the Court Lady Tamamo-no-ma} (1820) by Yashima
    Gakutei}
\end{figure*}
\clearpage 

\begin{figure*}
  \centering
   \includegraphics[width=\linewidth]{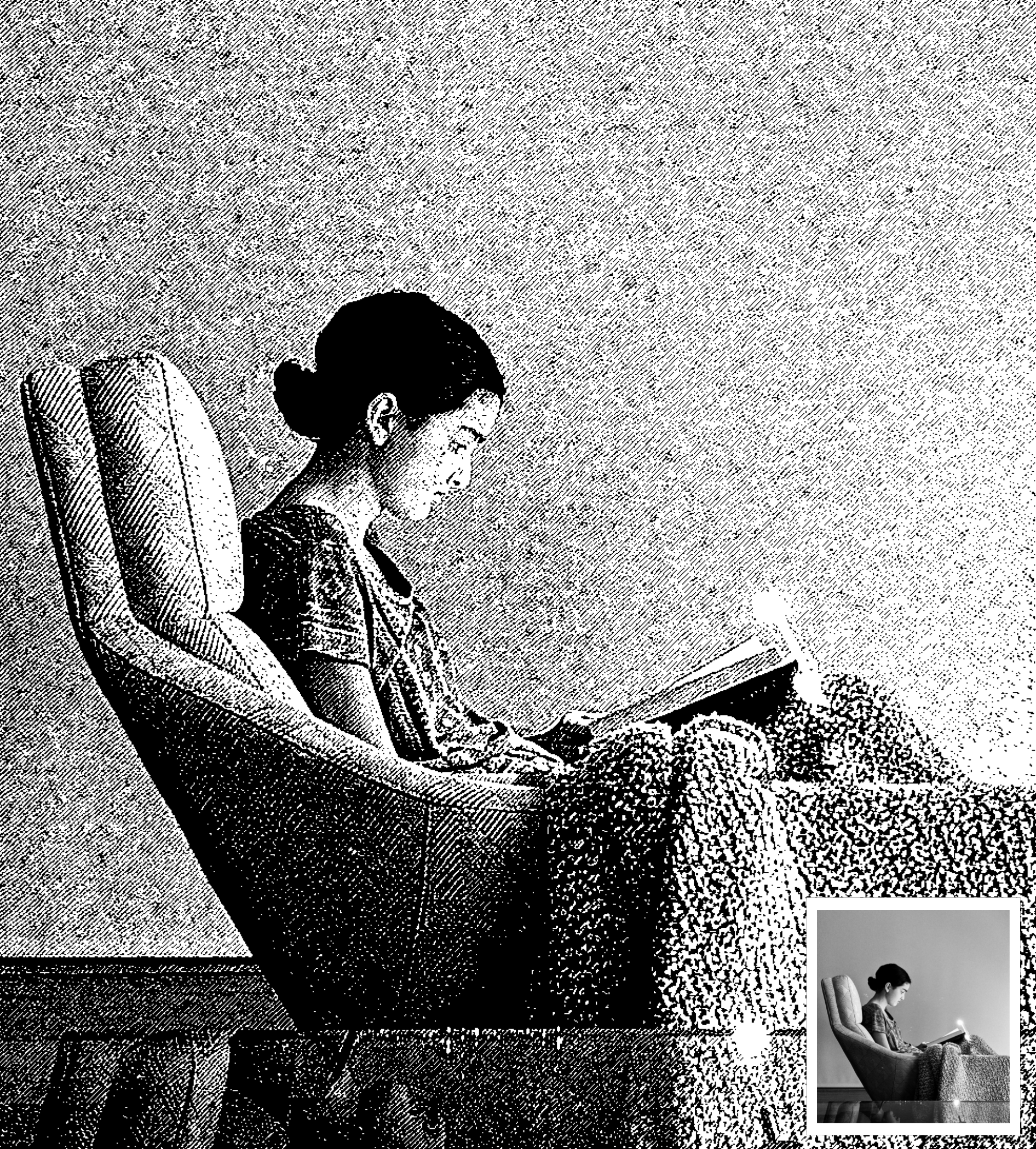}
  \caption{Black and white photograph of a young girl reading
    (2020)}
\end{figure*}
\clearpage

\begin{figure*}
\centering
  \includegraphics[width=\linewidth]{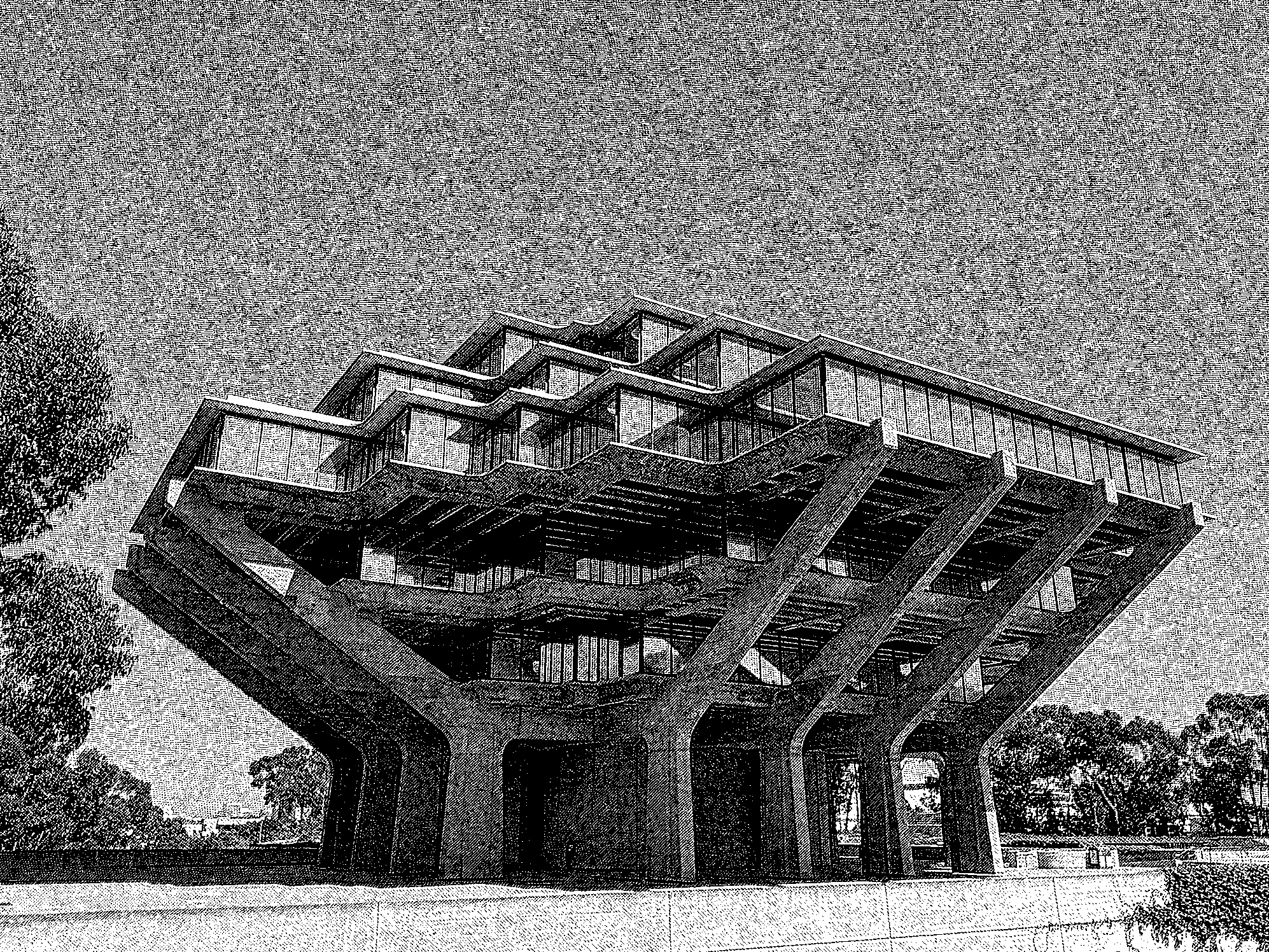}
  ~\\
  \includegraphics[width=0.3\linewidth]{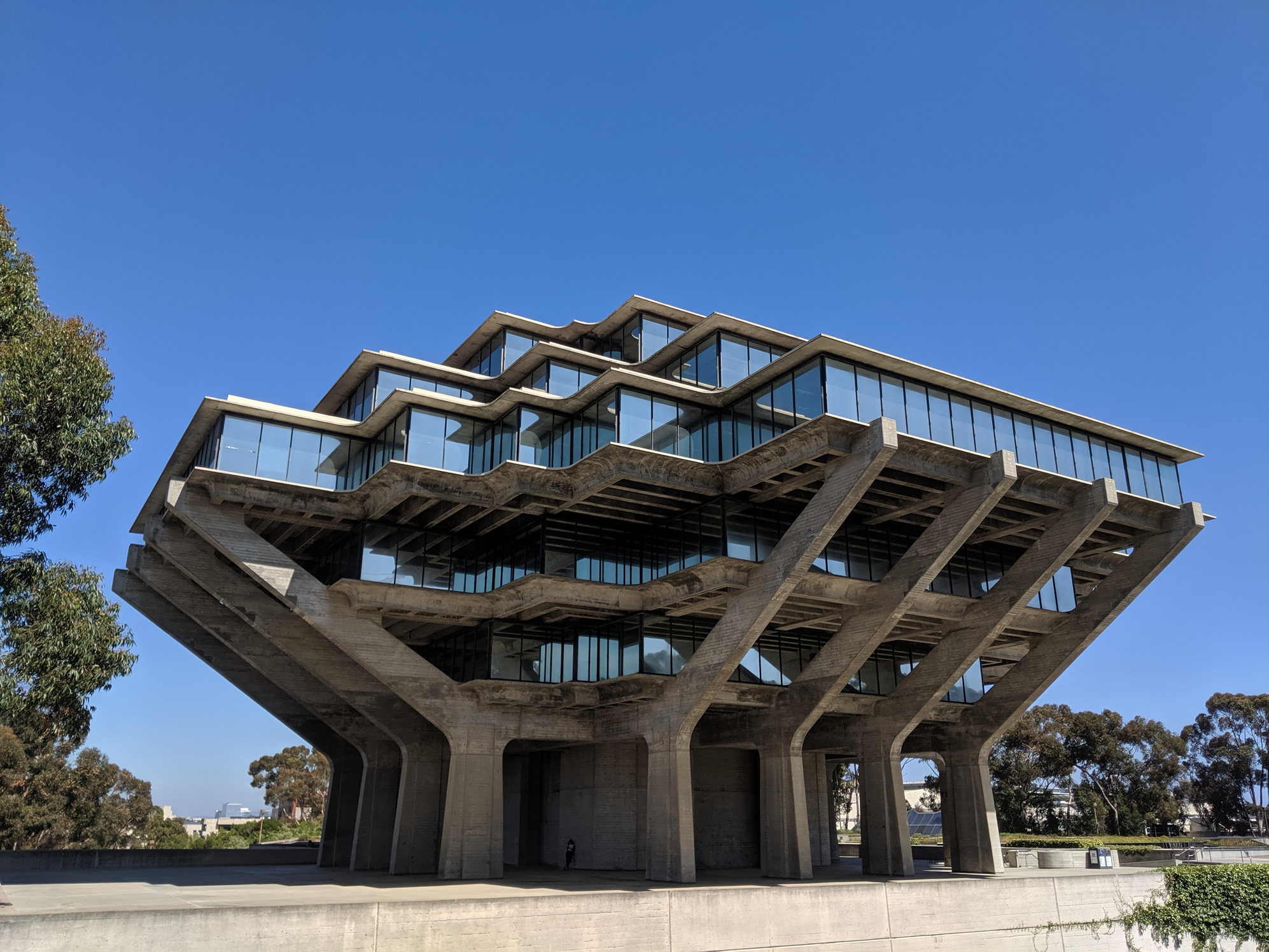}
  \caption{Color photograph (2021).}
\end{figure*}
\clearpage 

\begin{figure*}
  \centering
  \includegraphics[height=6in]{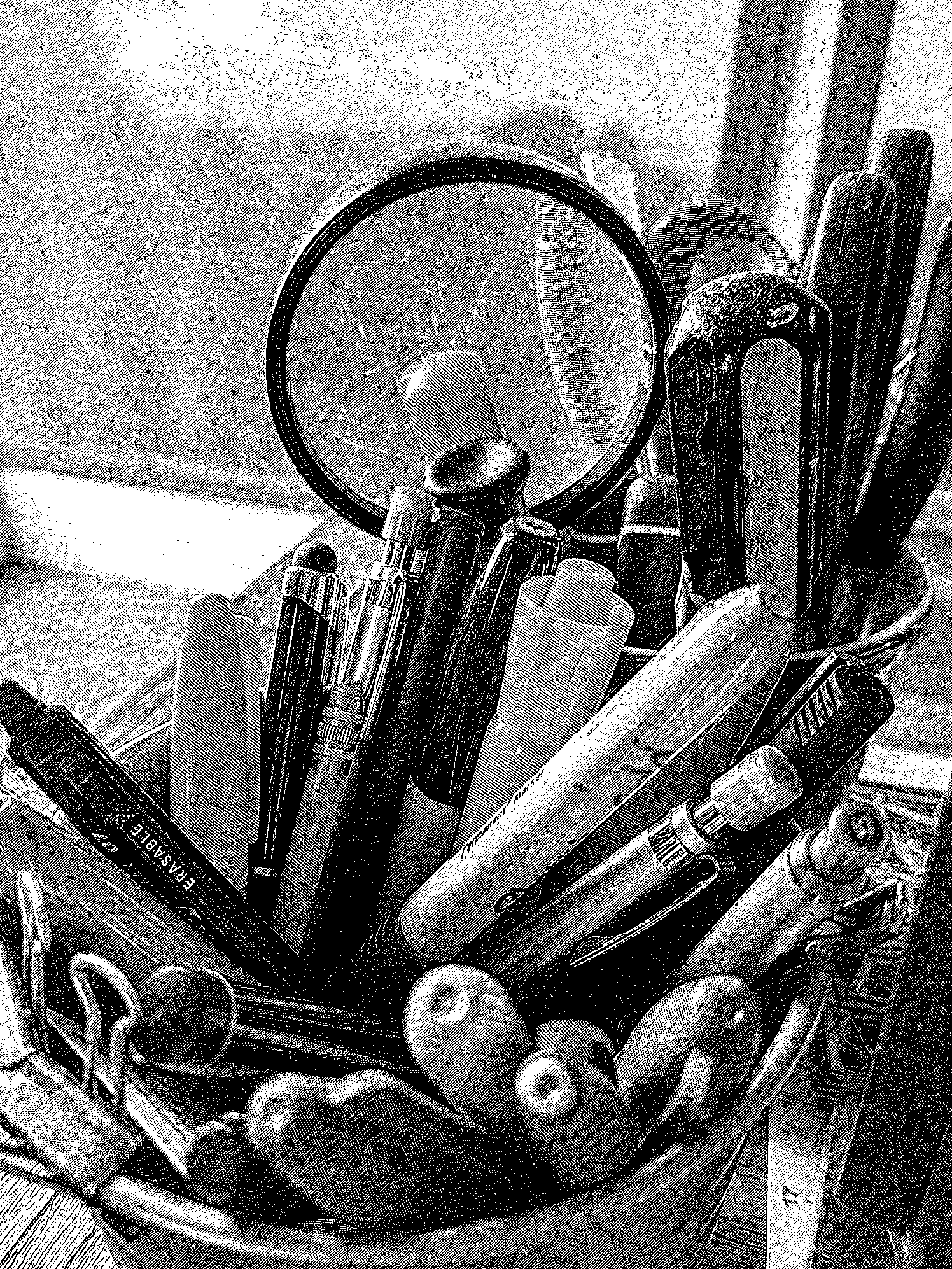}\\
  \includegraphics[height=1.8in]{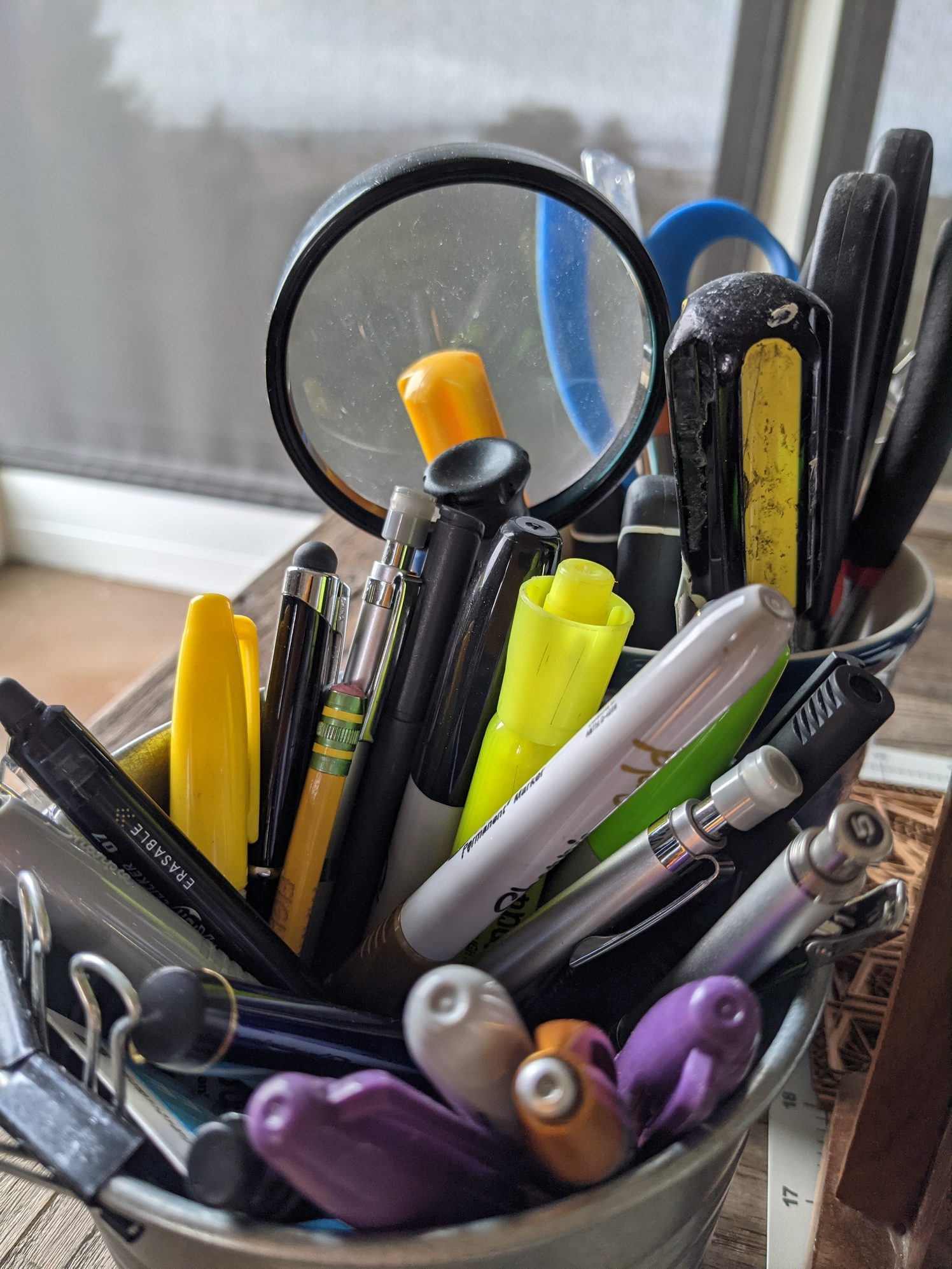}
  \caption{Color photograph (2021).}
\end{figure*}
\clearpage

\begin{figure*}
  \centering
  \includegraphics[width=\linewidth]{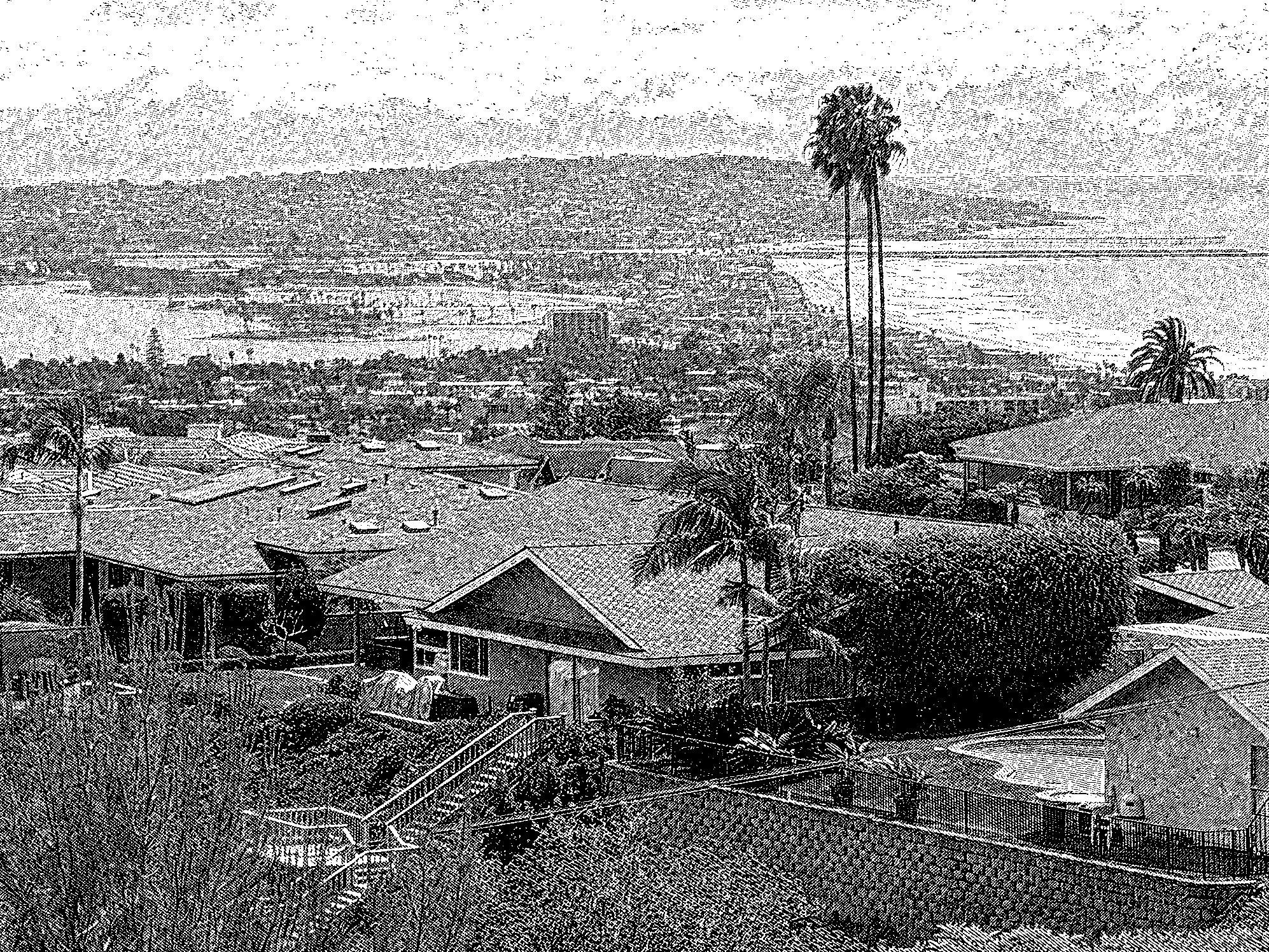}
  ~\\  
  \includegraphics[width=0.3\linewidth]{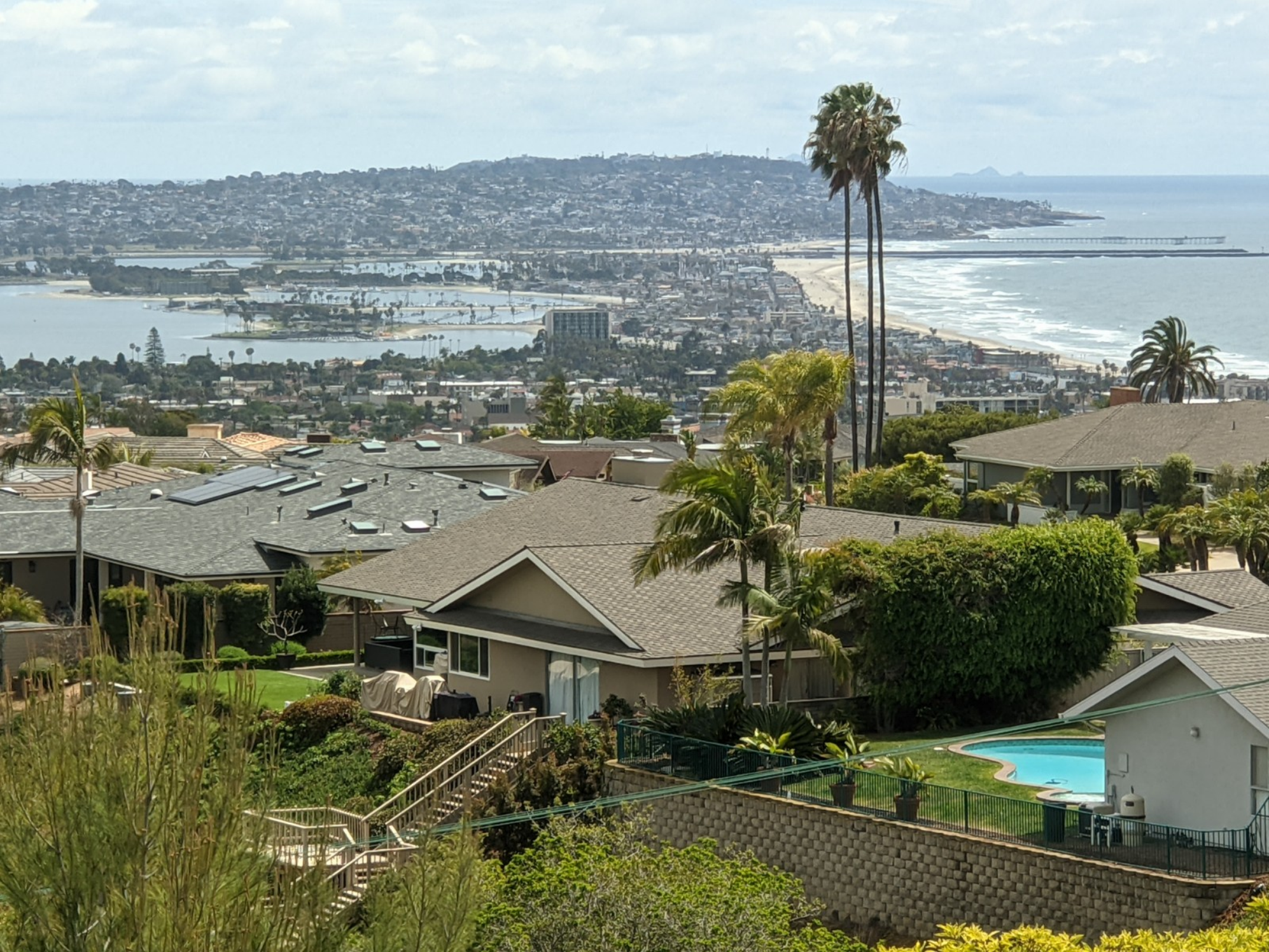}
  \caption{Color photograph (2021).}
\end{figure*}
\clearpage 

\subsection{Historic uses of Hatching and Cross-Hatching}
\begin{figure*}[h]
    \centering

  a. \includegraphics[height=2.1in]{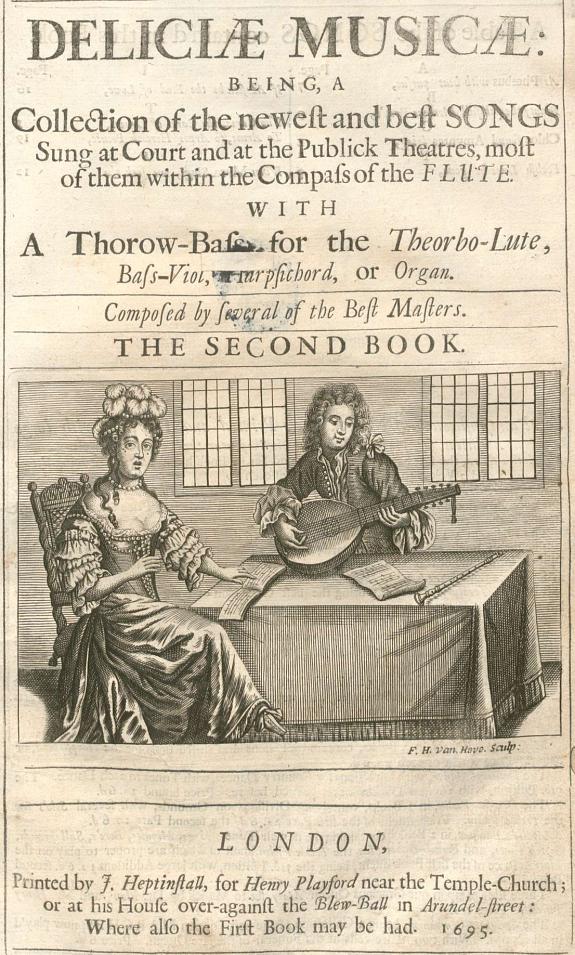} \:\: \:\: \:\: 
  b. \includegraphics[height=2.1in]{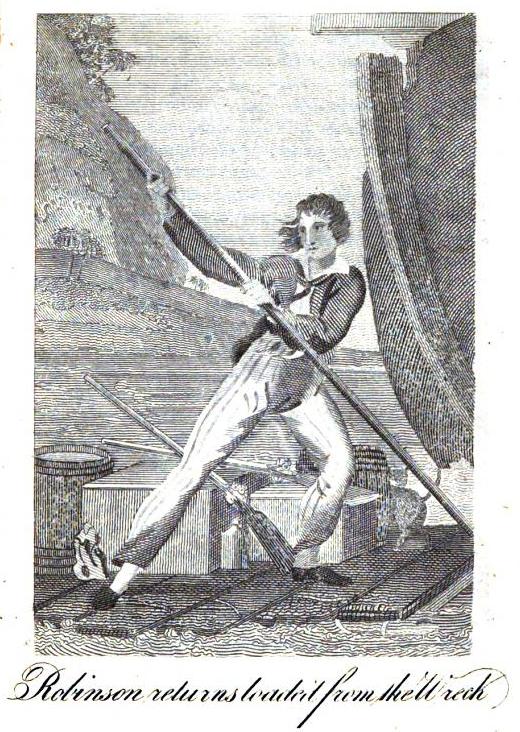}\\ \: \\
  c. \includegraphics[height=2.1in]{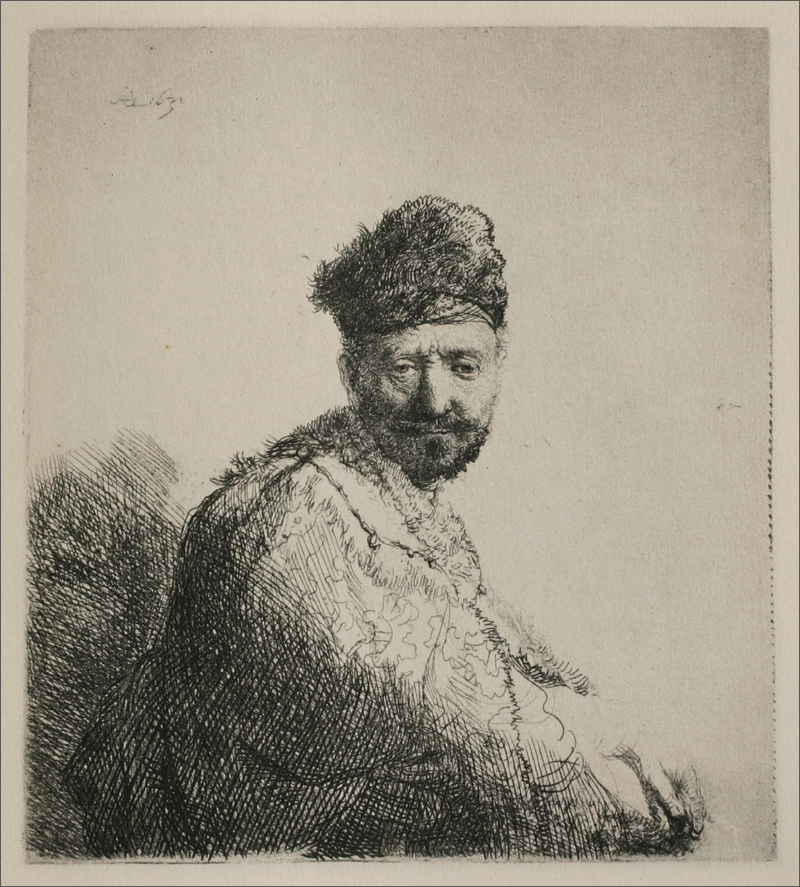}\\
    \caption{Three historic uses of hatching and cross-hatching.  (a)
      From 1695 book \emph{Deliciae Musicae: The second
      book}~\cite{1695deliciae} (b) from  Campe, J.H. (1825)~\cite{campe1825robinson} (c)
      \emph{A Man with a Short Beard and Embroidered Cloak} (1631) by
      Rembrandt.}
    \label{fig:hatching}
\end{figure*}

\end{document}